\documentclass[dvipsnames]{article} %
\PassOptionsToPackage{numbers,sort&compress}{natbib}
\usepackage{colm2024_conference}

\usepackage{booktabs}
\usepackage{graphicx}
\usepackage{enumitem}
\usepackage{wrapfig}
\usepackage{algorithm}
\usepackage{algpseudocode}

\usepackage{microtype}
\usepackage{amsmath}
\usepackage{colortbl}
\usepackage[utf8]{inputenc}
\usepackage[T1]{fontenc}
\usepackage[french,vietnamese,mongolian,greek,english]{babel}

\usepackage{caption}
\usepackage{subcaption}
\usepackage{setspace}
\usepackage{url}
\usepackage{multirow}
\usepackage{colortbl}
\usepackage{tabularx}
\usepackage{blindtext}
\usepackage{pgfplots}
\pgfplotsset{compat=1.18} 
\usepackage{tikz}
\usetikzlibrary{er,positioning,bayesnet}
\usepackage{makecell}
\usepackage{tipa}
\usepackage{siunitx}
\usepackage{nicefrac}
\usepackage{tocloft}
\usepackage{listings}
\usepackage[raster,skins]{tcolorbox} %
\usepackage{xltabular}
\usepackage{adjustbox}
\usepackage{xurl}
\usepackage{rotating}
\usepackage[normalem]{ulem}
\usepackage{chngcntr}
\useunder{\uline}{\ul}{}

\usepackage{pgffor}
\usepackage{xstring}

\usepackage{amsmath,amsfonts,bm}

\def\eqref#1{equation~\ref{#1}}

\def\1{\bm{1}}

\DeclareMathAlphabet{\mathsfit}{\encodingdefault}{\sfdefault}{m}{sl}
\SetMathAlphabet{\mathsfit}{bold}{\encodingdefault}{\sfdefault}{bx}{n}

\newcommand{\ie}{i.\,e., }
\newcommand{\eg}{e.\,g., }

\newcommand*\justify{%
  \fontdimen2\font=0.4em
  \fontdimen3\font=0.2em
  \fontdimen4\font=0.1em
  \fontdimen7\font=0.1em
  \hyphenchar\font=`\-
}

\renewcommand{\texttt}[1]{%
  \begingroup
  \ttfamily
  \begingroup\lccode`~=`/\lowercase{\endgroup\def~}{/\discretionary{}{}{}}%
  \begingroup\lccode`~=`[\lowercase{\endgroup\def~}{[\discretionary{}{}{}}%
  \begingroup\lccode`~=`.\lowercase{\endgroup\def~}{.\discretionary{}{}{}}%
  \catcode`/=\active\catcode`[=\active\catcode`.=\active
  \justify\scantokens{#1\noexpand}%
  \endgroup
}

\title{\raisebox{-0.4ex}{\includegraphics[height=1.2em]{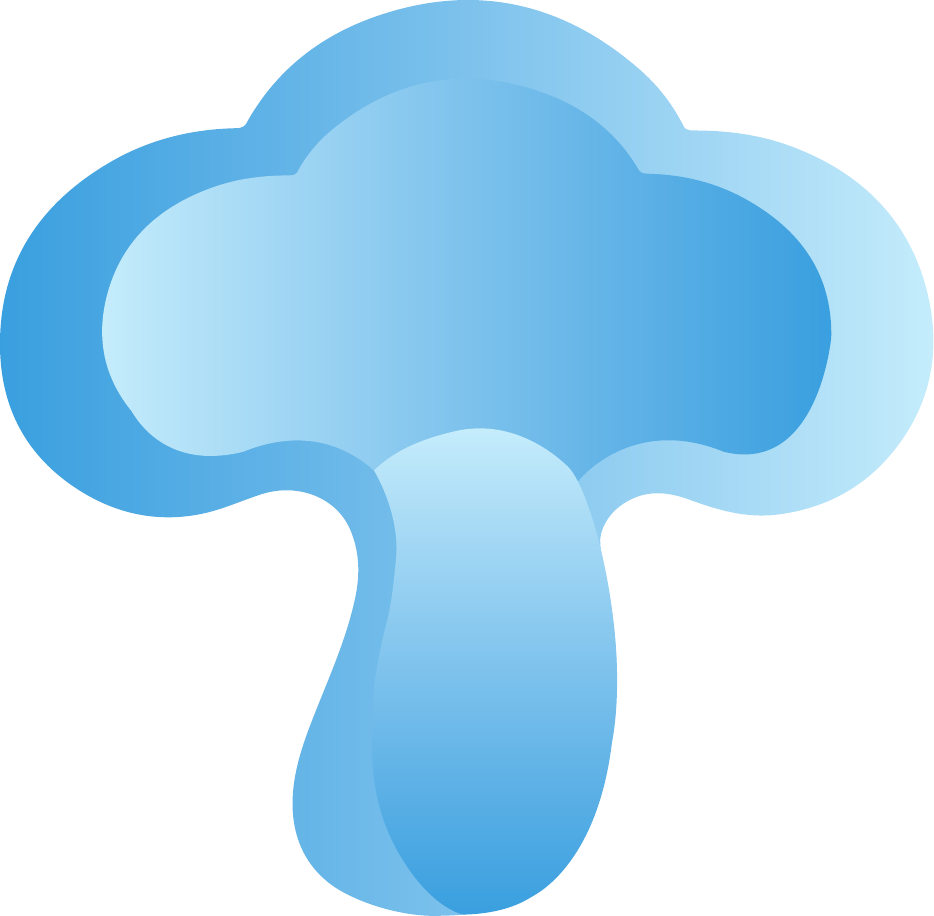}}  UMind-VL : A Generalist Ultrasound Vision-Language Model\\ for Unified Grounded Perception and Comprehensive Interpretation }

\author{Dengbo Chen, Ziwei Zhao, Kexin Zhang, Shishuang Zhao, Junjie Hou, Yaqian Wang\\ [5pt]
Nianxi Liao, Anlan Sun, Fei Gao, Jia Ding\textsuperscript{+}, Yuhang Liu\textsuperscript{*}, Dong Wang\textsuperscript{*}\\[9pt]
\bf Technical Report \\[6pt]
\bf Yizhun Medical AI Team\\
}

\begin{document}

\maketitle

\renewcommand{\thefootnote}{\fnsymbol{footnote}}
\footnotetext[2]{Corresponding Author: jia.ding@yizhun-ai.com}
\footnotetext[1]{Project Leads: \{yuhang.liu, dong.wang\}@yizhun-ai.com}

\vspace{-0.6cm}

\begin{abstract}

Despite significant strides in medical foundation models, the ultrasound domain lacks a comprehensive solution capable of bridging low-level Ultrasound Grounded Perception (e.g., segmentation, localization) and high-level Ultrasound Comprehensive Interpretation (e.g., diagnosis, reasoning). To bridge this gap, we propose UMind-VL, a unified foundation model designed to synergize pixel-level structural understanding with complex clinical reasoning. We first introduce UMind-DS, a large-scale multimodal dataset comprising 1.2 million ultrasound image–text pairs across 16 anatomical regions, enriching standard data with pixel-level annotations and clinician-validated rationales. Architecturally, UMind-VL incorporates a lightweight Dynamic Convolutional Mask Decoder that generates masks via dynamic kernels conditioned on LLM outputs. This design, combined with task-specific tokens, unifies segmentation, detection, geometric measurement, and diagnosis tasks within a single framework. Extensive evaluations demonstrate that UMind-VL significantly outperforms existing generalist multimodal models and achieves performance on par with, or superior to, state-of-the-art specialist models across segmentation, detection, keypoint localization, and diagnostic reasoning benchmarks, while maintaining strong generalization ability.
We demonstrate the capability of UMind-VL in Figure~\ref{fig:abs}.

\end{abstract}

\begin{figure}[h]
    \centering
    \includegraphics[width=0.85\textwidth]{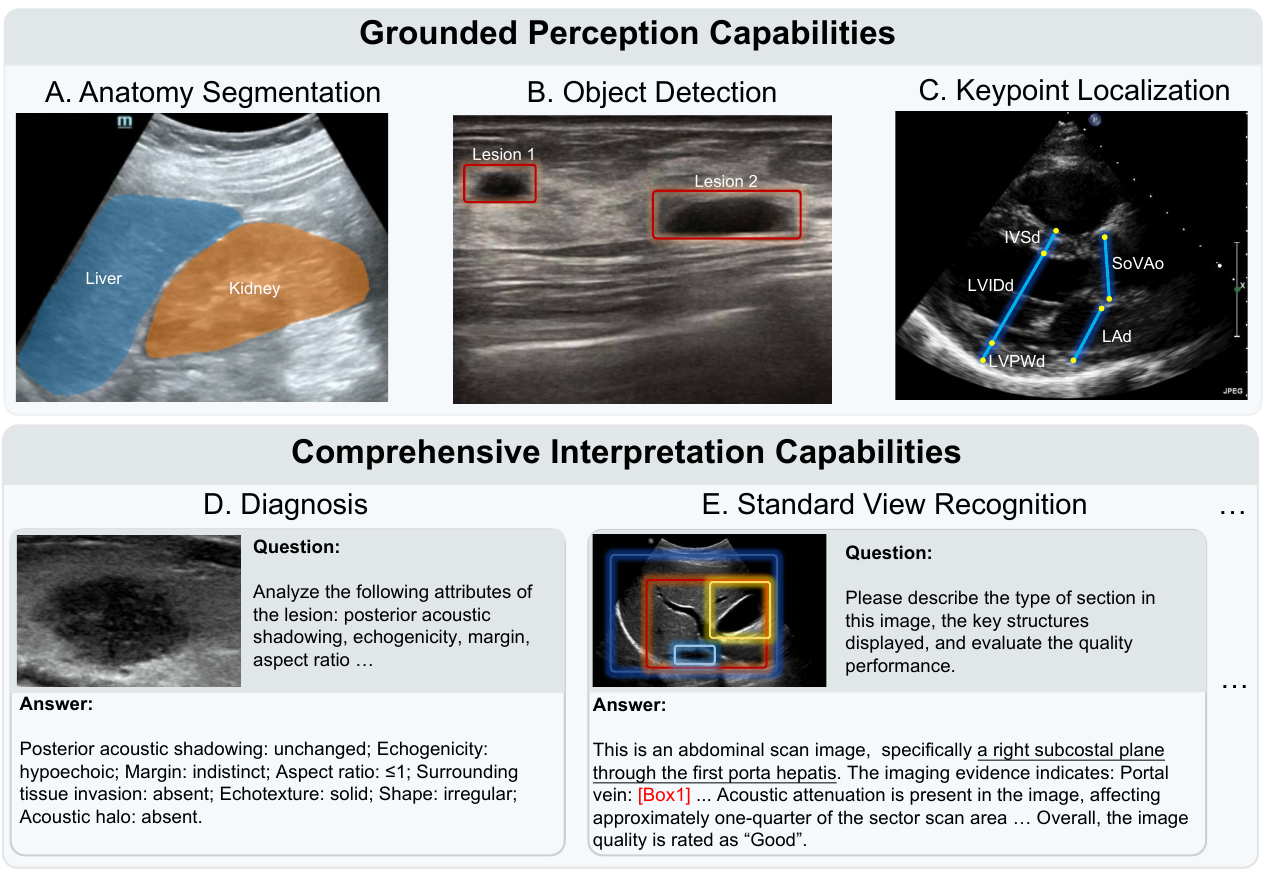}
    \caption{A snapshot of UMind-VL’s
 capabilities across a diverse range of ultrasound tasks.}
    \label{fig:abs}
\end{figure}

\newpage

\section{Introduction}

Remarkable advancements in large language models (LLMs) and multimodal large language models (MLLMs) have significantly accelerated the progress of artificial intelligence in healthcare. This surge has catalyzed the development of numerous specialized and general-purpose medical foundation models, such as BioGPT~\citep{luo2022biogpt}, Med-PaLM~\citep{singhal2023large, singhal2025toward}, and Baichuan-M2~\citep{dou2025baichuan}, each tailored for diverse clinical applications. Furthermore, multimodal systems like Lingshu~\citep{xu2025lingshu} and MedGemma~\citep{sellergren2025medgemma} have demonstrated strong capabilities in reasoning across heterogeneous medical data, including clinical text and various medical imaging modalities. Despite this rapid progress, the \textbf{Ultrasound} domain lacks a foundation model that is sufficiently comprehensive to meet the complex demands of real-world clinical workflows.

Clinical ultrasound tasks can be conceptually organized into a hierarchy ranging from low-level perception to high-level abstraction. The first level concerns the perception of fundamental visual structures—such as organ segmentation, keypoint detection, and lesion localization. We refer to this capability as \textbf{Ultrasound Grounded Perception}, emphasizing spatial grounding and structural understanding at the pixel or region level. The second level involves abstract, clinically oriented reasoning, including view classification, lesion characterization, and pathology prediction. We term this capability \textbf{Ultrasound Comprehensive Interpretation}, as it requires integrating visual semantics with domain knowledge to produce clinically meaningful insights. While Ultrasound Grounded Perception serves as the indispensable foundation for interpretability and reliability, the accuracy of Ultrasound Comprehensive Interpretation ultimately determines the clinical utility of a foundational model.

To develop a general-purpose ultrasound foundation model that is robust across the full spectrum of clinical tasks, these two levels must be seamlessly integrated. However, existing literature reveals a persistent disconnect between them (detailed in Section~\ref{sec:related_work}). Grounded Perception is predominantly approached through segmentation-based paradigms—such as adaptations of the Segment Anything Model (SAM)~\citep{kirillov2023segment}—which excel at delineating anatomical regions but offer limited diagnostic insight. Conversely, Comprehensive Interpretation is typically addressed using vision–language models (VLMs) that map visual inputs to labels or reports, often lacking explicit grounding in fine-grained anatomical contexts. This functional dichotomy prevents existing systems from achieving clinically coherent reasoning, where anatomical grounding and diagnostic inference operate synergistically.

\begin{figure}[hb]
\centering
\includegraphics[width=\textwidth]{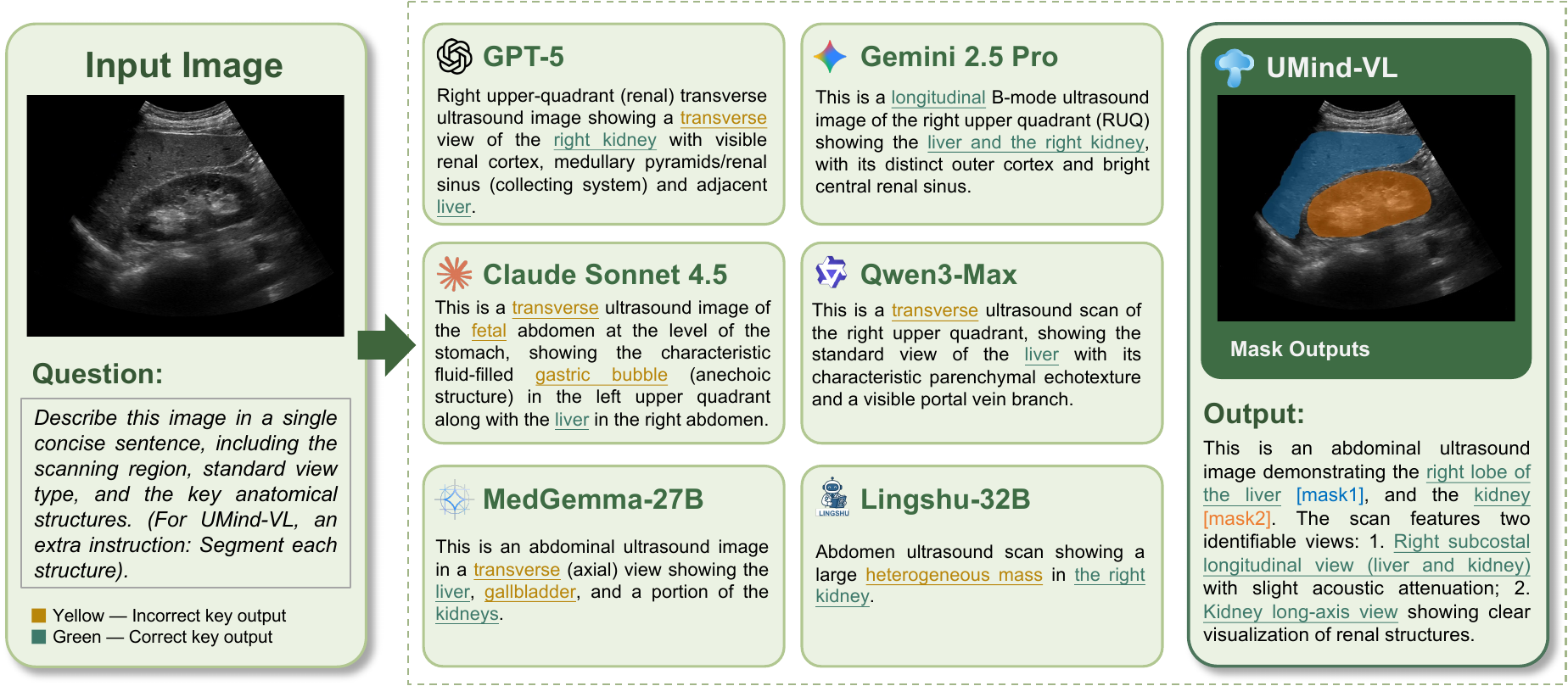}
\caption{\textbf{Comparative evaluation of LLMs for anatomical structure recognition.} Using a standard abdominal ultrasound view containing the liver and kidney, we observe that all public LLMs—except for Gemini 2.5 Pro—struggle to accurately identify all structures. UMind-VL not only correctly interprets the images but also provides pixel-level segmentation grounding for both organs.}
\label{fig:intro_cmp}
\end{figure}

Given the flexibility and generalization capacity of VLMs, integrating explicit Ultrasound Grounded Perception directly into these architectures offers a promising direction. However, it remains unclear whether existing VLMs inherently understand ultrasound anatomy. To investigate this, we evaluated several representative general and medical VLMs on typical clinical abdominal ultrasound images. As illustrated in Figure~\ref{fig:intro_cmp}, most models fail to accurately describe even fundamental anatomical structures, such as the liver and kidney. We attribute this deficiency to two main factors: (1) standard VLM architectures are not optimized for fine-grained spatial grounding; and (2) training data typically lack anatomically precise ultrasound vision–language alignments.

To bridge this divide, we propose UMind-VL, a unified foundation model explicitly designed for ultrasound clinical reasoning. Recognizing that Grounded Perception and Comprehensive Interpretation must operate synergistically, we adopt a co-design strategy spanning both data curation and architectural innovation.

On the data front, we introduce UMind-DS, a multimodal corpus of 1.2 million ultrasound image–text pairs covering 16 anatomical regions and diverse imaging modalities (2D, M-mode, and Doppler). This resource integrates 815k clinically validated real-world samples with 424k synthetically enhanced examples designed to reinforce anatomical grounding, mitigate hallucinations, and cover rare pathologies. Crucially, UMind-DS provides pixel-level structural annotations for organs and lesions, alongside clinician-validated diagnostic rationales, enabling the joint optimization of spatial semantics and clinical decision-making.

To endow UMind-VL with a robust understanding of visual structures, we introduce targeted architectural adaptations. A critical bottleneck in current general-purpose VLMs is the absence of pixel-level grounding—a prerequisite for precise medical image interpretation. We address this by incorporating a lightweight \textbf{Dynamic Convolutional Mask Decoder}. Unlike heavy segmentation heads, this efficient design generates masks via dynamic kernels conditioned on LLM outputs, implicitly driving the multimodal backbone to capture fine-grained spatial features. Furthermore, to accommodate diverse clinical tasks, we unify segmentation, detection, and geometric measurement (\eg points and lines) through a set of task-specific special tokens. This framework not only streamlines the architecture but also fosters a deep synergy between Grounded Perception and Comprehensive Interpretation.

Our extensive experimental validation covers segmentation, detection, keypoint localization, and diagnosis tasks across varying anatomical systems. Results show that UMind-VL consistently outperforms existing generalist and medical multimodal models. More importantly, it achieves performance on par with or superior to state-of-the-art specialist models (e.g., surpassing Mask2Former in segmentation and rivaling Deform-DETR in detection), demonstrating robust generalization even on out-of-distribution datasets. Furthermore, through extensive qualitative visualizations, we showcase UMind-VL’s versatility across a broad spectrum of ultrasound tasks, highlighting its remarkable generalization capabilities in complex clinical scenarios.

Our main contribution can be summarized as follows:

\begin{enumerate}
\item We propose UMind-VL, the first ultrasound foundation model to unify Grounded Perception and Comprehensive Interpretation, enabling simultaneous precise localization and detailed medical reasoning within a single framework.
\item We construct UMind-DS, a substantial and comprehensive ultrasound dataset containing 1.2 million image-text pairs with extensive coverage of anatomical regions and imaging protocols.
\item Extensive experiments demonstrate that UMind-VL consistently surpasses existing generalist multimodal models and matches or exceeds state-of-the-art specialist models across various ultrasound tasks.It exhibits strong robustness and generalization on out-of-distribution datasets, and qualitative analyses further validate its versatility across diverse real-world ultrasound scenarios.
\end{enumerate}

\section{Related Work}
\label{sec:related_work}

\subsection{Large Language Model for Medicine}
Large language models (LLMs) have revolutionized medical AI by unlocking advanced capabilities in processing unstructured clinical text. Pioneering domain-specific models like BioGPT~\citep{luo2022biogpt} (trained on biomedical literature) and GatorTron~\citep{yang2022gatortron} (trained on electronic health records) laid the groundwork for specialized medical foundation models. Subsequent advances yielded general-purpose systems such as Med-PaLM~\citep{singhal2023large} and Med-PaLM 2~\citep{singhal2025toward}, which achieve expert-level performance on medical exams, alongside clinically adapted models like Baichuan-M2~\citep{dou2025baichuan} and HuatuoGPT~\citep{chen2024huatuogpto1, zhang2023huatuogpt}.

The integration of vision capabilities further catalyzed progress, giving rise to medical multimodal large language models (MLLMs). Systems like Lingshu~\citep{xu2025lingshu} and MedGemma~\citep{sellergren2025medgemma} enable joint reasoning over heterogeneous data (e.g., imaging and clinical notes), while adaptations such as LLaVA-Med~\citep{li2023llava}, HuatuoGPT-Vision~\citep{chen2024huatuogpt}, and radiology-focused RadFM~\citep{wu2025towards} align visual features from X-rays/CTs with diagnostic semantics. These MLLMs excel at tasks like visual question answering and report generation, yet none address the full clinical workflow demands of ultrasound. Meanwhile, current ultrasound foundation models (e.g., Dolphin-V1~\citep{weng2025dolphin}, EchoCare~\citep{zhang2025fully}) remain limited in unifying fine-grained anatomical grounding with high-level diagnostic reasoning—a critical gap for real-world deployment.

\subsection{Unified Visual Perception in Medical Multimodal Models}
Clinically viable ultrasound foundation models require robust Ultrasound Grounded Perception—the ability to anchor high-level reasoning to precise anatomical structures through spatial reasoning (e.g., organ segmentation, lesion localization, keypoint detection for anatomical landmarks). Current medical multimodal models exhibit fragmented grounding capabilities across two paradigms. Segmentation-Centric approaches like MedSAM~\citep{ma2024segment} and UltraSAM~\citep{meyer2025ultrasam}, built upon the Segment Anything Model (SAM)~\citep{kirillov2023segment}, achieve high-fidelity delineation of organs and lesions. However, they function as isolated perception modules that lack diagnostic semantics and cannot interface with reasoning components. Vision-language models (VLMs) like the paradigms of LLaVA-Med~\citep{li2023llava} and RadFM~\citep{wu2025towards} inherit coarse anatomical grounding capabilities (e.g., bounding boxes) from general multimodal LLMs. While excelling at Comprehensive Interpretation (e.g., pathology classification), they lack pixel-level precision for ultrasound-critical tasks like cardiac valve motion tracking or fetal biometric measurement.

General MLLMs attempt to bridge this gap through several strategies.  Embedding SAM~\citep{kirillov2023segment} or dedicated mask decoders into MLLM backbones~\citep{lai2024lisa,wu2024visionllm,rasheed2024glamm,ren2024pixellm} enhances segmentation fidelity while introducing architectural complexity and impeding end-to-end training. Approximating masks via polygons~\citep{wang2023visionllm,wang2024git} or textual labels~\citep{lantext4seg,wang2024git} sacrifices anatomical accuracy for tokenizer compatibility, which is unacceptable in precision-sensitive ultrasound applications. UFO~\citep{tang2025ufo} reframes segmentation as an embedding retrieval task: a learnable mask token embedding computes dot-product similarity with dense image features, and high-similarity spatial locations are aggregated to reconstruct the segmentation mask. This formulation offers a more balanced integration of fine-grained perception and semantic reasoning—though room for improvement remains.

\section{Methodology}

\subsection{Overall Architecture}

\begin{figure}[t]
    \centering
    \includegraphics[width=\textwidth]{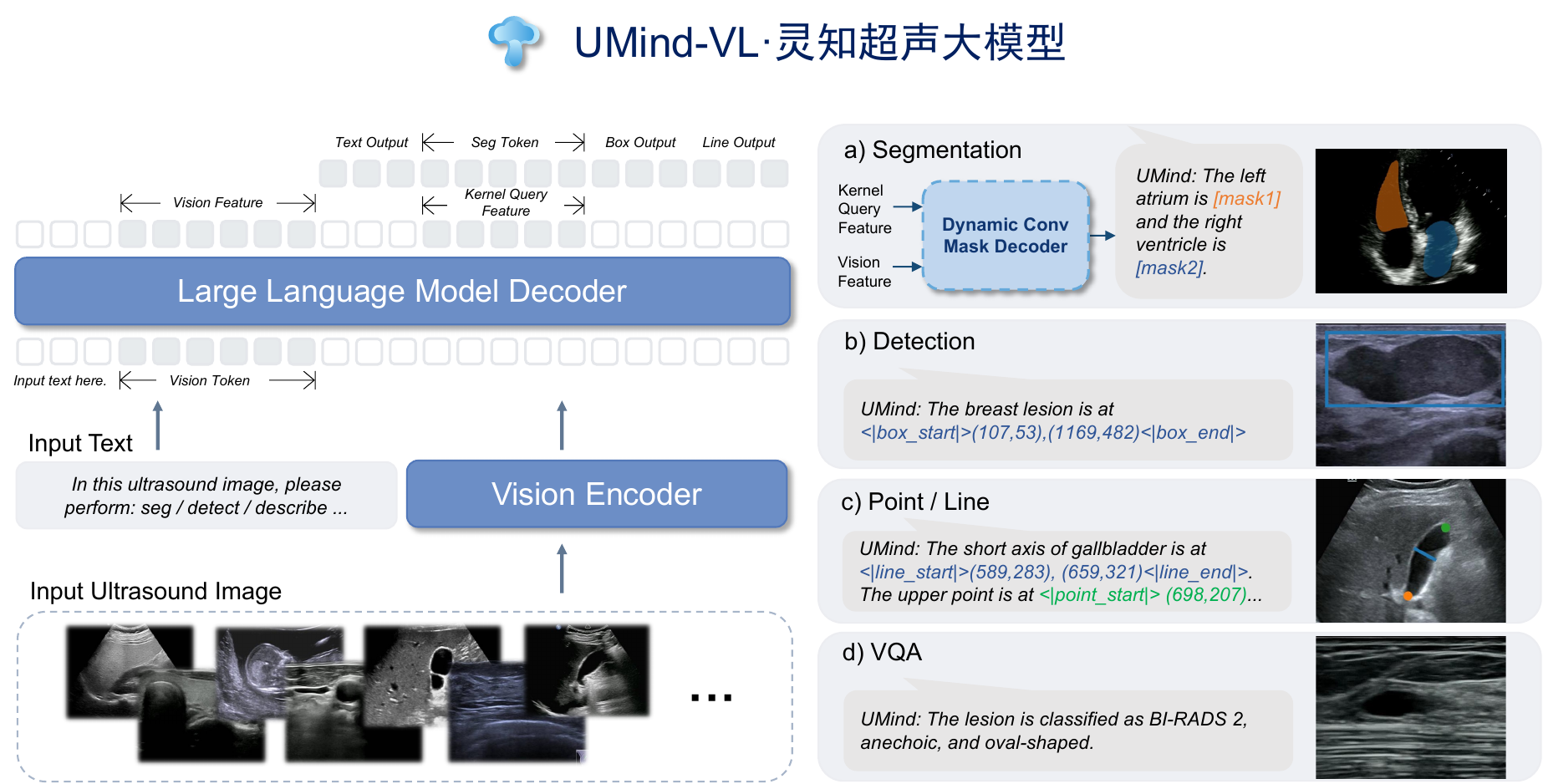}
    \caption{\textbf{Overall architecture and functions of UMind-VL.} UMind-VL performs detection, point/line prediction, and VQA directly via token prediction, while the segmentation function employs an additional lightweight mask decoder. Note that mask identifiers (e.g., \texttt{[mask1]}) in the visualization denote the aggregation of $N$ specially designed \texttt{<|seg\_mask|>} tokens.}
    \label{fig:main_arch}
\end{figure}

The architecture of UMind-VL is designed to unify Ultrasound Grounded Perception and Ultrasound Comprehensive Interpretation within a single framework. Regarding model outputs, Ultrasound Grounded Perception tasks encompass mask segmentation, bounding-box detection, and point or line prediction, whereas Comprehensive Interpretation typically manifests through visual question answering (VQA). As illustrated in Figure~\ref{fig:main_arch}, UMind-VL follows an encoder--decoder paradigm, comprising a vision encoder for ultrasound feature extraction and a large language model (LLM) decoder for task-conditioned multimodal reasoning and token prediction.

Specifically, the vision encoder processes input ultrasound images to extract a sequence of vision tokens. Simultaneously, free-form text inputs specifying the desired task are tokenized into text tokens. Both sequences are subsequently fed into the LLM decoder.

A pivotal innovation of UMind-VL is its unified output interface for ultrasound tasks. Inspired by \citet{tang2025ufo}, we accommodate diverse tasks by leveraging the LLM decoder to predict flexible and adaptable token sequences. For the majority of tasks—including detection, point/line prediction, and VQA—outputs are generated directly within this sequence. As shown in Figure~\ref{fig:main_arch} (b-d), spatial grounding for detection and point/line tasks is achieved using special tokens to demarcate coordinates.

Conversely, segmentation requires a distinct mechanism to produce high-resolution, pixel-level outputs. As depicted in Figure~\ref{fig:main_arch} (a), UMind-VL addresses this by introducing a Dynamic Conv Mask Decoder. Unlike methods relying on external pretrained segmentation models (e.g., bridging VLM with SAM), our design philosophy aims to intrinsically activate the spatial grounding capability of the VLM itself. Consequently, the Dynamic Conv Mask Decoder adopts a relatively lightweight design, compelling the vision encoder and LLM decoder to learn the necessary grounding abilities during end-to-end training.

This unified output interface implicitly enforces a shared feature representation space across all tasks. This crucial design promotes synergy, enabling joint learning across diverse vision-language tasks to yield mutually reinforcing performance gains. We elaborate on these designs in the following sections.

\subsection{Ultrasound Grounded Perception Representation}

To support a unified and expressive grounding interface tailored for ultrasound imagery, UMind-VL extends the coordinate tokenization scheme of the Qwen-VL series~\citep{bai2025qwen2, wang2024qwen2} by introducing a richer set of structured spatial tokens. These tokens facilitate fine-grained control over geometric elements commonly encountered in clinical ultrasound tasks, enabling the model to reason over boxes, points, lines, and masks within a single vocabulary space.

\textbf{Bounding-box representation.}
Adhering to the Qwen-VL design, we adopt the \texttt{<|box\_start|>} $(x_1,y_1),(x_2,y_2)$ \texttt{<|box\_end|>} format to represent detected anatomical structures or lesions. This convention allows the model to localize regions using consistent syntax, facilitating seamless adaptation from general VLM grounding to ultrasound-specific detection.

\textbf{Keypoint representation.}
Ultrasound examinations frequently require precise localization of anatomical landmarks (e.g., the left ventricular endocardium in echocardiography for calculating the left ventricular ejection fraction (LVEF), a critical metric for heart function). To accommodate such fine-grained tasks, we introduce a dedicated point token format: \texttt{<|point\_start|>} $(x,y)$ \texttt{<|point\_end|>}. Compared to bounding boxes, the point representation provides a more compact and precise supervision signal, enabling the model to capture subtle structural cues essential for clinical workflows.

\textbf{Line representation.}
For tasks involving organ measurement, line segments serve as critical geometric primitives. We therefore design a line token format, \texttt{<|line\_start|>} $(x_1,y_1),(x_2,y_2)$ \texttt{<|line\_end|>}, allowing the LLM decoder to explicitly model linear relationships between anatomical points. This representation supports downstream measurement reasoning and enhances the model’s understanding of elongated ultrasound structures.

\textbf{Segmentation representation.}
Unlike detection or keypoint prediction, segmentation necessitates dense pixel-level outputs that cannot be fully expressed by coordinate tokens alone. To unify segmentation within the token-based interface, we introduce a special token, \texttt{<|seg\_mask|>}, where the corresponding LLM feature is directly consumed by the Dynamic Conv Mask Decoder (refer to Section~\ref{sec:mask_decoder}). During training, the LLM learns to emit a sequence of $N$ \texttt{<|seg\_mask|>} tokens when segmentation is required, and the decoder maps the associated feature embeddings into a high-resolution binary mask. This design tightly couples VLM reasoning with spatial grounding, allowing segmentation quality to improve jointly with the model’s overall multimodal understanding.

\textbf{Object-reference mechanism.}
Maintaining consistency across multimodal reasoning steps is crucial for complex ultrasound tasks. Consistent with Qwen-VL, we incorporate \texttt{<|object\_ref\_start|>} and \texttt{<|object\_ref\_end|>} to enclose predicted spatial primitives—boxes, points, lines, and masks—when they are referenced by subsequent tokens. This mechanism enables coherent cross-token referencing, empowering the LLM to link spatial outputs with semantic descriptions.

Collectively, these structured grounding tokens form a coherent and expressive instruction space for ultrasound perception tasks. By integrating geometric primitives, object references, and mask-specific tokens into a unified tokenization framework, UMind-VL enables the LLM to seamlessly mix spatial grounding with high-level reasoning, ultimately supporting a broad spectrum of clinical ultrasound applications.

\subsection{Dynamic Conv Mask Decoder}
\label{sec:mask_decoder}

To equip UMind-VL with dense pixel-level segmentation capabilities while preserving a lightweight and tightly coupled architecture, we introduce a Dynamic Convolutional Mask Decoder. This module transforms LLM-produced mask-query embeddings into high-resolution masks.

For segmentation tasks, UMind-VL is trained to emit a sequence of $N$ consecutive \texttt{<|seg\_mask|>} tokens for each target. The LLM decoder features associated with these tokens are extracted to serve as \textbf{Kernel Query Features} (see Figure~\ref{fig:main_arch}), acting as conditioning signals for generating dynamic convolution kernels. Simultaneously, the LLM outputs vision features corresponding to the input ultrasound image tokens, which serve as the second input to the mask decoder.

\begin{figure}[t]
    \centering
    \includegraphics[width=\textwidth]{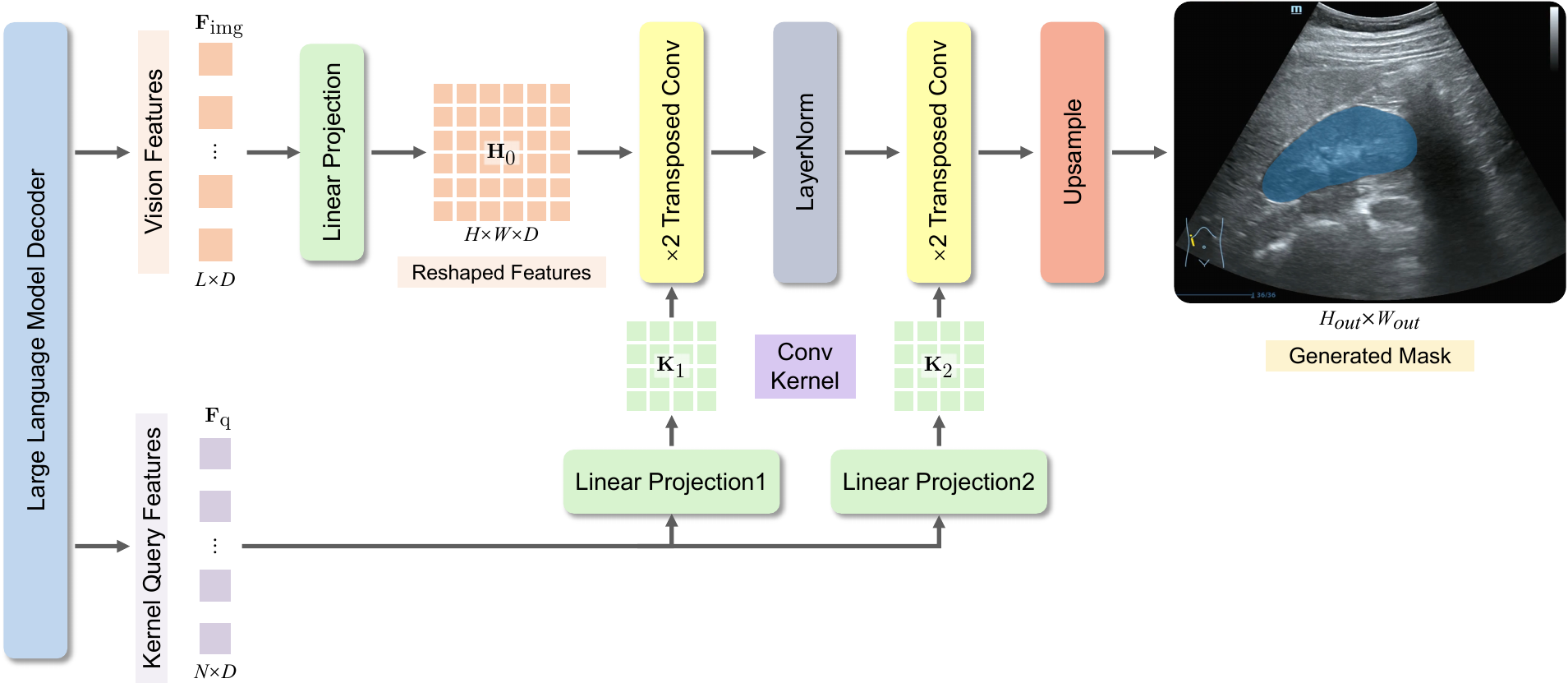}
    \caption{\textbf{Dynamic Convolutional Mask Decoder.}}
    \label{fig:mask_decoder}
\end{figure}

\paragraph{Input feature preparation.}
As illustrated in Figure~\ref{fig:mask_decoder}, the mask decoder consumes both Kernel Query Features and vision features. Let
\[
\mathbf{F}_{\text{img}} \in \mathbb{R}^{L \times D}
\]
denote the vision features produced by the LLM decoder, where $L = H \times W$ corresponds to the number of visual tokens and $D$ is the LLM feature dimension (with $H$ and $W$ determined by the downsampling ratio of the vision encoder). 

We first apply a learned token-wise linear projection to adjust the channel dimension:
\[
\mathbf{F}_{\text{img}}' = \mathrm{Proj}_{\text{img}}(\mathbf{F}_{\text{img}}) \in \mathbb{R}^{L \times C}.
\]
The projected sequence is then reshaped into a spatial feature map:
\[
\mathbf{H}_0 = \operatorname{reshape}(\mathbf{F}_{\text{img}}') \in \mathbb{R}^{H \times W \times C}.
\]

For each segmentation target, the $N$ emitted Kernel Query Features form a query matrix
\[
\mathbf{F}_{\text{q}} \in \mathbb{R}^{N \times D}.
\]
These features are linearly projected to generate the weights for two stages of dynamic transposed convolutions:
\[
\mathbf{K}_1 = \mathrm{Proj}_{\text{ker}}^{(1)}(\mathbf{F}_{\text{q}}), \qquad
\mathbf{K}_2 = \mathrm{Proj}_{\text{ker}}^{(2)}(\mathbf{F}_{\text{q}}),
\]
where $\mathbf{K}_1$ and $\mathbf{K}_2$ are reshaped into convolution kernels for subsequent upsampling.

\paragraph{Dynamic convolutional upsampling.}
The mask decoder progressively upsamples $\mathbf{H}_0$ into a high-resolution mask.  
First, a $2\times$ dynamic transposed convolution conditioned on $\mathbf{K}_1$ is applied:
\[
\mathbf{H}_1 = \mathrm{DeConv}_{\mathbf{K}_1}^{\times 2}(\mathbf{H}_0),
\]
followed by Layer Normalization:
\[
\mathbf{H}_1' = \mathrm{LayerNorm}(\mathbf{H}_1).
\]

A second $2\times$ dynamic transposed convolution further increases the spatial resolution:
\[
\mathbf{H}_2 = \mathrm{DeConv}_{\mathbf{K}_2}^{\times 2}(\mathbf{H}_1').
\]

Finally, an upsampling operator (e.g., bilinear interpolation) refines the output to the target resolution:
\[
\hat{\mathbf{M}} = \mathrm{Upsample}(\mathbf{H}_2),
\]
yielding the final predicted mask logits
\[
\hat{\mathbf{M}} \in \mathbb{R}^{H_{\text{out}} \times W_{\text{out}}}.
\]

\paragraph{Design motivation and advantages.}
Beyond offering a unified interface for ultrasound segmentation, the proposed Dynamic Conv Mask Decoder provides the following advantages:

\textbf{(1) Lightweight architecture.}
The decoder introduces only three linear layers and a LayerNorm module as learnable parameters. This lightweight design maintains acceptable computational overhead while ensuring sufficient expressive power through dynamically generated convolutional kernels.

\textbf{(2) Encouraging grounding capability within the VLM backbone.}
By intentionally constraining the decoder's capacity, the majority of the learning burden is shifted toward the VLM backbone. This design choice incentivizes the model to internalize stronger pixel-level grounding abilities rather than relying on a heavy task-specific head. Such enhanced grounding capacity directly benefits broader Comprehensive Interpretation tasks by promoting anatomically consistent and semantically aligned predictions.

\subsection{Optimization Objective}
\label{sec:optimization}

UMind-VL is optimized using a unified objective that integrates both language modeling and dense prediction supervision. For samples lacking segmentation annotations, the model is trained in the same manner as standard vision-language models, \ie by minimizing the autoregressive token prediction loss over the entire output sequence.

For training samples that include segmentation masks, we introduce an additional supervision branch for the Dynamic Conv Mask Decoder. The predicted masks are optimized using a weighted combination of region-overlap and pixel-level classification losses, ensuring stable convergence across diverse lesion shapes and scales. This auxiliary segmentation objective is jointly optimized with the token-level loss in an end-to-end fashion, allowing the model to simultaneously learn semantic reasoning and fine-grained grounding without the need for task-specific training stages.

\section{Data Curation}

\begin{figure}[h]
    \centering
    \includegraphics[width=\textwidth]{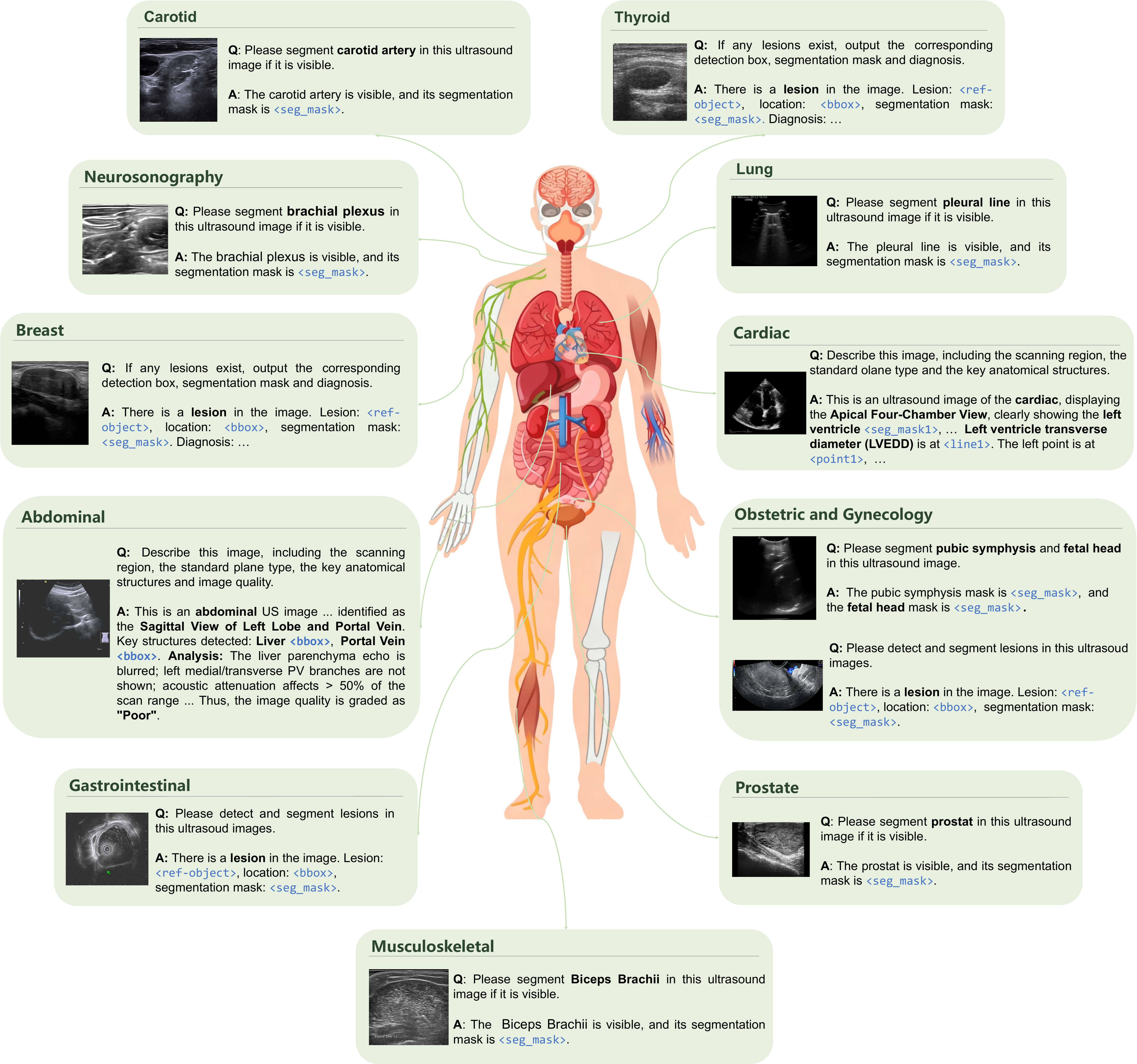}
    \caption{\textbf{Representative examples across major anatomical regions in UMind-DS.} Each panel illustrates a VQA-style instruction–response pair for a specific organ system, demonstrating the dataset’s coverage of segmentation, detection, lesion characterization, anatomical identification, and other clinically relevant tasks. The illustrative anatomical figure is generated using Seedream 4.0~\citep{seedream2025seedream}.}
    \label{fig:body}
\end{figure}

This section details the collection and synthesis of the multimodal dataset \textbf{UMind-DS}, which serves as the foundation for developing UMind-VL ultrasound model. Our objective is to construct a comprehensive and unified data resource that enables robust learning across a broad spectrum of downstream tasks, including segmentation, detection, key point localization, and visual question answering (VQA). 
Figure~\ref{fig:body} provides representative examples of ultrasound data across major anatomical regions, illustrating the diversity of organ systems and task types included in UMind-DS.

To achieve sufficient diversity and scale, we curate a wide range of open-source datasets from publicly available repositories and construct a large-scale in-house dataset that spans multiple organs, imaging protocols, and clinical conditions. These datasets collectively ensure a balanced representation of anatomical structures and pathological variations, establishing the foundation for subsequent post-training and instruction tuning.

Beyond the acquisition of real-world data, we incorporate data synthesis to enhance specific model capabilities. Synthetic samples are generated to improve instruction-following behavior, mitigate hallucinations, and augment the coverage of underrepresented question–answer types. This combination of large-scale real data and controlled synthetic data ensures both breadth of generalization and depth of clinical understanding in the UMind-VL model.

\begin{figure}[htpb]
    \centering
    \begin{subfigure}{0.53\linewidth}
        \centering
        \includegraphics[width=\linewidth]{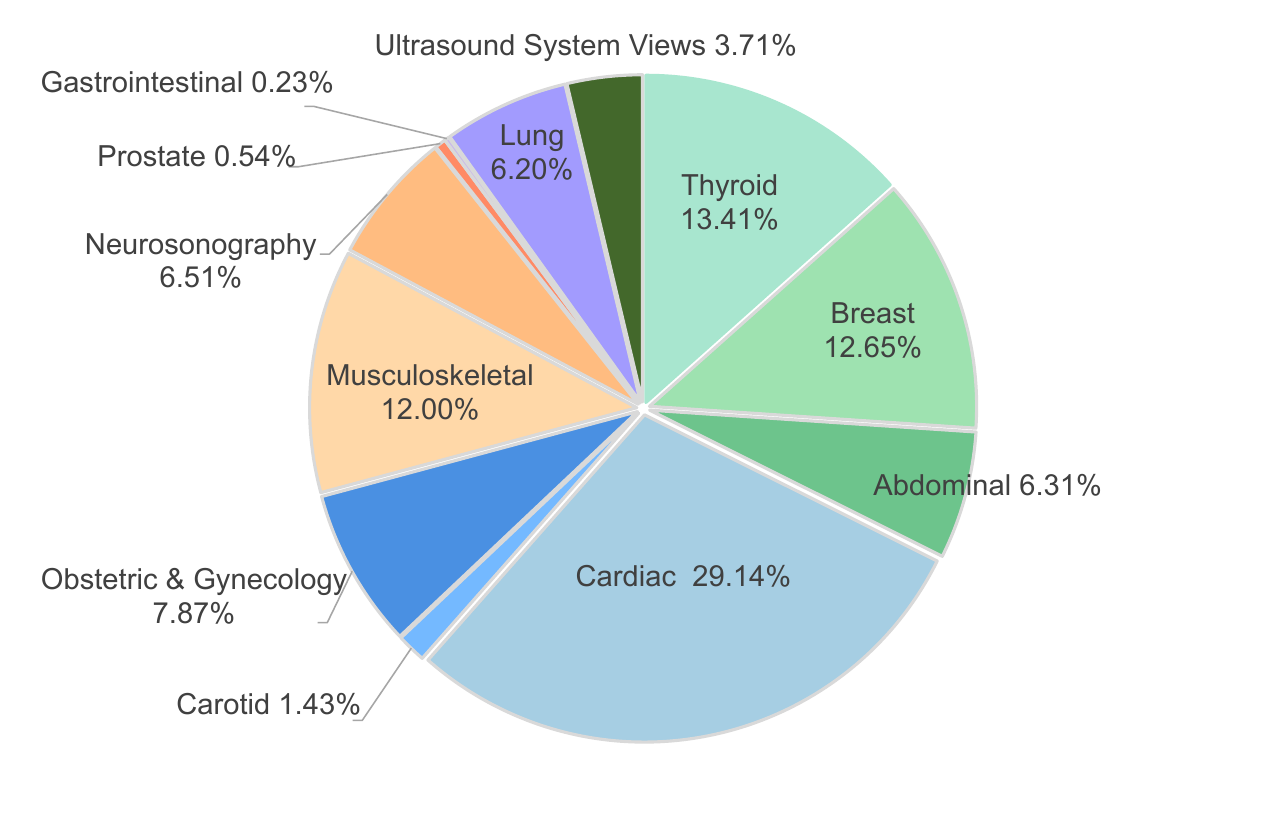}
        \caption{Distribution across anatomical categories.}
        \label{fig:sub1}
    \end{subfigure}
    \hfill
    \begin{subfigure}{0.41\linewidth}
        \centering
        \includegraphics[width=\linewidth]{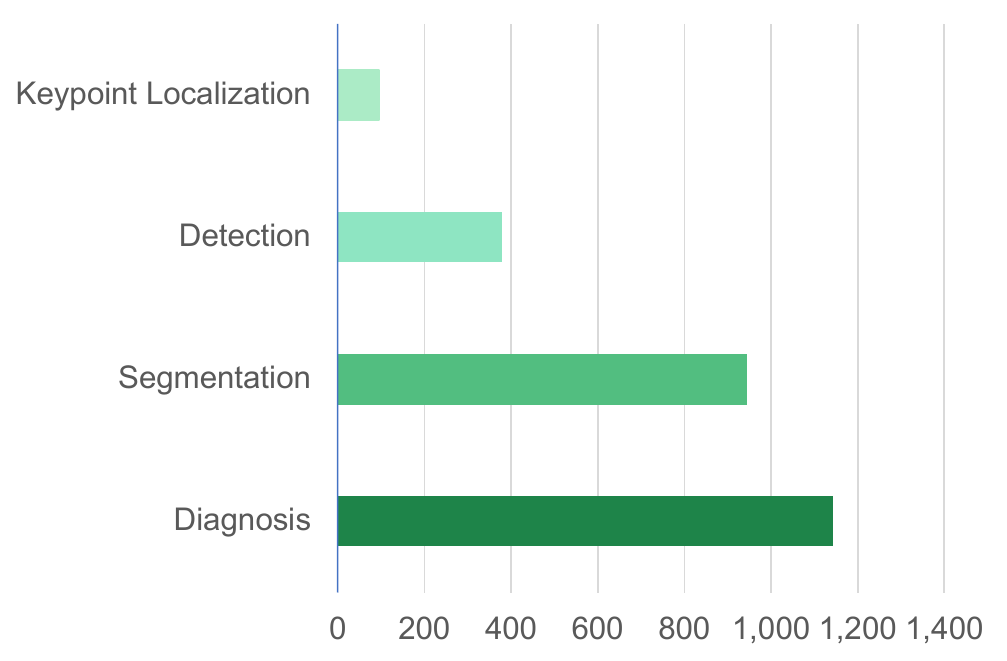}
        \caption{Task-wise composition (in thousands).}
        \label{fig:sub2}
    \end{subfigure}
    \caption{\textbf{Overview of the UMind-DS composition.} (a) The anatomical distribution of UMind-DS, illustrating coverage across major clinical categories including thyroid, breast, abdominal, cardiac, carotid, obstetric and gynecology, musculoskeletal, neurosonography, prostate, gastrointestinal, lung and system views.
(b) Task-wise composition of the collected samples (in thousands), including segmentation, detection, keypoint localization, and diagnostic question answering. Each sample may involve multiple tasks, and the statistics represent the number of samples that include each respective task. }
    \label{fig:data_distribution}
\end{figure}

\subsection{Data Collection}
We first describe the motivation, strategy, and workflow for collecting and harmonizing diverse datasets from both public and in-house sources.
Figure~\ref{fig:data_distribution} provides an overview of the UMind-DS composition, illustrating its anatomical coverage across major clinical categories as well as the task-wise distribution of collected samples.
Table \ref{tab:medical_datasets} enumerates all collected public datasets, categorized by organ system or content type.

\begin{table}[h]
\centering
\caption{List of all collected public datasets.}
\label{tab:medical_datasets}
\begin{tabular}{p{4.5cm} p{10.5cm}}
\toprule
\textbf{Type of Data} & \textbf{Collected Public Datasets} \\
\midrule
Thyroid & AUITD~\citep{AUITD}, KFGNet~\citep{wang2022key}, DDTI~\citep{pedraza2015open}, TG3K~\citep{gong2021multi}, Thyroid\_Ultrasound\_Cine\_Clip~\citep{thyroid_cine_2021}, TN3K~\citep{gong2021multi}, TN5k~\citep{zhang2025tn5000}\\
\midrule
Breast & BUSIS~\citep{zhang2022busis}, BUV~\citep{10.1007/978-3-031-16437-8_59}, GDPH-SYSUCC~\citep{10015121}, BrEaST~\citep{pawlowska2024curated}, BUID~\citep{homayoun2022applications, hamyoon2022artificial, ardakani2023open}, BUS\_UC~\citep{iqbal2023busuc}, BUS-UCLM~\citep{vallez2024busuclm}, BUS-BRA~\citep{gomez2024bus}, BUS\_DatasetB~\citep{yap2017automated}, BUSI ~\citep{al2020dataset},
S1~\citep{guo2021segmentation}, STU-Hospital~\citep{stuhospital2020}, BUS-COT~\citep{yu2025chain}\\
\midrule
Abdominal & Fatty-Liver~\citep{byra2018transfer}, C-TRUS~\citep{leenings2024c}, LEPset~\citep{li2023dsmt}, USAnotAI~\citep{USAnotAI}, 105US~\citep{dataset}, AbdomenUS~\citep{orlando2020ussimandsegm}, AUL~\citep{xu_yiming_2022_7272660}, kidneyUS~\citep{singla2023open}\\
\midrule
Cardiac & MVSEG2023~\citep{carnahan2023segmentation}, MEIS~\citep{tseng2024real}, CAMUS~\citep{leclerc2019deep}, CardiacUDC~\citep{xu2021cardiacudc}, EchoCP~\citep{leclerc2019deep}, EchoNet-Dynamic~\citep{ouyang2020video}, EchoNet-Pediatric~\citep{reddy2023video}, UnityImaging~\citep{duffy2022high, huang2022fix, unity2022}\\
\midrule
Carotid & CUBS~\citep{meiburger2021carotid}, MI-SegNet~\citep{bi2023mi}, CCAUI~\citep{momot2022cca}\\
\midrule
Obstetric and Gynecology & FETAL\_PLANES\_DB~\citep{burgosfetal_planes_db}, FPUS23~\citep{prabakaran2023fpus23}, African-Fetal-Standard-Plane~\citep{sendra2023maternal}, PCOSGEN~\citep{divekar2024leveraging}, FUGC~\citep{FUGC}, uterine-fibroid-ultrasound~\citep{uterine-fibroid-ultrasound}, ACOUSLIC~\citep{SAPPIA2025103640}, FASS~\citep{correggio2023fetal}, Fast-U-Net~\citep{ashkani2022fast}, FH-PS-AOP~\citep{jieyun_2024_10829116}, HC~\citep{van2018automated}, MMOTU-2D~\citep{DBLP:journals/corr/abs-2207-06799}, MMOTU-3D~\citep{DBLP:journals/corr/abs-2207-06799}\\
\midrule
Musculoskeletal & Leg-3D-US~\citep{gonzalez2024methods}, ASUS~\citep{ungi2020automatic}, FALLMUD~\citep{cunningham2018estimating, cronin2020automated, michard2021aw}, STMUS\_NDA~\citep{marzola2021transverse}\\
\midrule
Neurosonography & Brachial\_Plexus~\citep{tyagi2024nerve}, 
Segthy~\citep{kronke2022tracked}, UBPD~\citep{ding2022mallesnet}, Ultrasound\_Nerve\_Segmentation~\citep{montoya2016ultrasound})\\
\midrule 
Prostate & MicroSeg~\citep{shao2024micro}, RegPro~\citep{baum2023mr}\\
\midrule
Gastrointestinal & Appendix~\citep{marcinkevivcs2023regensburg}, GIST514-DB~\citep{he2022query2}\\
\midrule
Lung & COVID-BLUES~\citep{wiedemann2025covid}, Lung-Disease-Classification~\citep{katumba2025dataset}, LUSS\_Phantom~\citep{mclaughlan2024lung}\\
\midrule
Natural Images & LLM-Seg40K~\citep{Wang_2024_CVPR}, RefCOCO~\citep{kazemzadeh2014referitgame}\\
\bottomrule
\end{tabular}
\end{table}

\subsubsection{Segmentation Data Collection}
Segmentation data are curated to support dense anatomical perception and lesion delineation across a broad spectrum of clinical scenarios. The dataset includes ultrasound images from multiple organs and physiological systems, covering thyroidal~\citep{pedraza2015open,kronke2022tracked,gong2021multi,zhang2025tn5000},
breast~\citep{pawlowska2024curated, ardakani2023open, vallez2024busuclm, gomez2024bus, iqbal2023busuc}, 
cardiac~\citep{tseng2024real,leclerc2019deep,xu2021cardiacudc,ouyang2020video,reddy2023video}, 
carotid~\citep{meiburger2021carotid,bi2023mi,momot2022cca},
hepatic~\citep{byra2018transfer}, 
biliary~\citep{orlando2020ussimandsegm}, 
pancreatic~\citep{li2023dsmt}, 
renal~\citep{singla2023open}, 
splenic~\citep{USAnotAI},  
prenatal~\citep{van2018automated}, 
cervical~\citep{FUGC},  
prostatic~\citep{shao2024micro}, 
gastrointestinal~\citep{he2022query2} regions, etc~\citep{montoya2016ultrasound,ungi2020automatic}.

Despite this breadth, several key organs—notably the thyroid, breast, liver, pancreas, kidney, and heart—remain constrained by the limited size, diversity, and annotation rigor of publicly available ultrasound segmentation datasets. To address these gaps, we establish a proprietary, high-quality segmentation in-house dataset through a multi-stage annotation workflow. Each study is first annotated by a certified sonographer following standardized clinical protocols.
Then, all masks undergo double-blind review by two senior radiologists.
When disagreements arise, a senior clinical expert acts as arbiter to reach the final decision.
To further enhance the model's multimodal capacity, we additionally incorporate parts of a large-scale natural image segmentation dataset~\citep{Wang_2024_CVPR} to encode natural-image priors and ultrasound-specific semantics.

\subsubsection{Detection Data Collection}
Detection datasets are curated to support lesion-level and organ-level object detection. 
However, the availability and quality of public ultrasound detection datasets remain limited~\citep{zhang2025tn5000, yu2025chain}, presenting several notable challenges. 
First, public datasets are small-scale (e.g., TN5k with fewer than 5k annotated images) or suffer from inconsistent annotation. 
Moreover, some datasets concentrate on narrow clinical conditions or single-pathology cohorts, resulting in limited diversity in lesion types and imaging environments.
To overcome these limitations, we construct a large number of collected in-house datasets to ensure sufficient diversity and annotation reliability.
The dataset primarily covers the thyroid, breast, and abdomen regions, encompassing a broad spectrum of benign and malignant lesions.
Two experienced radiologists independently annotate every potential lesion in ultrasound videos in a double-blind manner, and discrepancies are adjudicated by a senior expert.
For each annotated lesion, the frame showing the most prominent appearance within its temporal span is manually selected as the representative image for the detection dataset.
Additionally, we integrate natural image object detection corpora~\citep{kazemzadeh2014referitgame} into the final detection dataset, enabling the model to benefit from both generic object priors and domain-specific lesion patterns.

\subsubsection{Keypoint Data Collection}
Keypoint localization datasets are curated for measurement and quantitative analysis tasks in ultrasound imaging. 
A key example is echocardiography, where accurate identification and measurement of the left ventricular ejection fraction (LVEF) are crucial for assessing heart function, diagnosing heart failure, and guiding interventions~\citep{gottdiener2004american}. Keypoint localization in this context involves identifying and marking the position of the left ventricular endocardium boundaries within the ultrasound image.
The datasets cover organs such as heart, liver, gallbladder, and kidney, integrating both public benchmarks~\citep{duffy2022high, huang2022fix, unity2022} and in-house annotated datasets.
Annotations are performed by trained sonographers, and the dataset is rigorously validated for metric reproducibility to ensure that model predictions align with established clinical measurement protocols, such as  LVEF assessment.

\subsubsection{VQA Data Collection}
The VQA datasets enable multimodal reasoning, allowing the model to answer clinically relevant questions about ultrasound images. It encompasses view recognition, anatomical identification, modality identification, diagnostic reasoning, and lesion characterization tasks.

We begin with public ultrasound VQA datasets~\citep{burgosfetal_planes_db,wiedemann2025covid} as foundational resources, then substantially expand them with private expert-curated QA datasets derived from real-world ultrasound annotations. These additions introduce clinically grounded questions that are rarely addressed in existing datasets, such as lesion benignity,  BIRADS scores~\citep{liberman2002breast}, and TIRADS scores~\citep{grant2015thyroid}, etc.

For comprehensive interpretation tasks such as lesion diagnosis and view classification, we introduce chain-of-thought (CoT) annotations~\citep{wei2022chain}. 
Annotators provide a step-by-step rationale. They begin with a holistic description of the ultrasound image, including the anatomical structures visible (e.g., liver, thyroid, breast). Then they note detailed observations of any lesions, such as location, size, shape, margin, echogenicity, and posterior acoustic features. The structured reasoning culminates in a final diagnostic conclusion or view classification.

By explicitly modeling the clinical decision-making process, CoT annotations not only guide large models to learn from medically grounded reasoning pathways but also enhance the interpretability of their predictions. This transparency is crucial for clinical adoption, as it enables practitioners to understand and validate the model's design-making process, thereby fostering trust in real-world medical applications.


\subsubsection{Data Cleaning and Standardization Pipeline}
To ensure data integrity, reliability, and consistency across heterogeneous ultrasound datasets, we established a unified data cleaning and standardization pipeline. 

First, an image quality control stage is applied to remove frames exhibiting motion blur, low contrast, or excessive noise, thereby preventing degraded supervision and ensuring visual consistency across imaging devices. 
Second, to mitigate redundancy and data leakage, duplicate samples are detected and subsequently pruned while maintaining diversity in organ types and pathological patterns. 
Third, incomplete metadata, such as missing organ labels, view types, or lesion attributes, are automatically completed through a hybrid approach combining task-specific models and consensus voting by MLLMs~\citep{chen2024internvl, bai2025qwen2, openai2025gptoss120bgptoss20bmodel,liu2024llavanext,liu2023improvedllava,liu2023llava}. Only predictions with high calibrated confidence are retained, while low-confidence samples are excluded to minimize label noise and maintain annotation reliability. 
Finally, all data were anonymized to remove patient identifiers, institutional information, and any embedded textual overlays, ensuring complete de-identification prior to downstream use.

\subsection{Data Synthesis}
To enhance instruction-following capabilities and reduce hallucination, we introduce synthetic data generation strategies that bridge gaps between tasks and improve model robustness.

\subsubsection{Instructional Data Synthesis}
In clinical practice and interactive applications, users often issue underspecified or abstract queries without explicit task definitions, such as “What can you observe in this scan?” or “Summarize the findings of this ultrasound.”
To enable the model to handle such open-ended inputs, we construct synthetic instructional data that simulate diverse interaction patterns and corresponding structured responses.

Each synthetic sample pairs a natural-language instruction with a hierarchically organized response, including organ recognition, anatomical structure description, lesion localization summaries, and diagnostic impressions.
For example, when prompted with “Please analyze this breast ultrasound image.”, the generated target response enumerates visible organs, describes their morphological boundaries, highlights any detected abnormalities, and concludes with a concise diagnostic statement.
These instructional samples encourage the model to reason systematically across perception and semantics while maintaining factual alignment between visual and textual modalities.
By incorporating both generic and task-specific templates, this synthesis step effectively enhances the model’s multi-task generalization and natural language comprehension capabilities.

\subsubsection{Alignment Data Synthesis}
Alignment data synthesis ensures that the model responds consistently and responsibly to ambiguous or clinically invalid queries. 
Medical image interpretation requires strict anatomical logic. For example, questions such as “Please segment the left ventricle in this image” may arise even if the image is not a cardiac ultrasound and the left ventricle is not present, representing a biologically inconsistent or out-of-scope query.

To address such risks, we generate counterfactual and adversarial question–answer pairs that explicitly represent ill-posed or contradictory scenarios.
During training, the model learns to identify out-of-scope, anatomically inconsistent, or logically impossible prompts and respond with context-aware rejection statements (e.g., “The question is not applicable to this anatomical region” or "This query is not applicable to the current region of interest").
This alignment data serves as a crucial safeguard, ensuring factual accuracy, anatomical consistency, and clinical reliability under diverse interaction conditions.
Together, the instructional and alignment synthesis stages substantially improve the robustness, interpretability, and trustworthiness of the UMind foundation model.

\subsection{Dataset Summary}
UMind-DS integrates heterogeneous data sources across multiple modalities, tasks, and organs.
It contains approximately \textbf{1.2} million text–image pairs curated for multimodal visual question answering and clinical reasoning, spanning \textbf{16} distinct organs and physiological systems. 
Among them, around 815k samples are derived from real-world clinical data, while 424k are generated through synthesis pipelines.
UMind-DS includes various ultrasound modalities, such as 2D, M-mode, PW, CW, and TDI, ensuring comprehensive coverage of clinical imaging scenarios.
It is designed to support both low-level perception tasks (segmentation, detection, keypoint localization) and high-level cognitive tasks (diagnosis, COT-based inference).
This comprehensive resource forms the foundation for post-training and instruction-tuning our multimodal ultrasound model, enabling robust performance across diverse clinical imaging scenarios.

\section{Experiments}
In this section, we first detail the implementation and training protocols of UMind-VL. Subsequently, we conduct a comprehensive evaluation, comparing our model against state-of-the-art generalist and specialist models across a series of critical clinical tasks.

\subsection{Training Details}
We build UMind-VL upon the Qwen3-VL-4B architecture, a vision-language foundation model with 4 billion parameters pretrained on extensive web-scale data. To bridge the gap between general vision and medical diagnostics, we perform Supervised Fine-Tuning (SFT) utilizing the curated UMind-DS dataset, which comprises 1.2 million ultrasound image–text pairs and covers 16 distinct organs and physiological systems.

For optimization, we fine-tune the model for 5 epochs employing low-rank adaptation (LoRA) to ensure parameter-efficient adaptation. Regarding architectural settings, the Dynamic Conv Mask Decoder is configured with $N=16$ and $C=128$. The training process is parallelized across a cluster of 128 GPUs using the AdamW optimizer. We implement a cosine learning rate schedule peaking at $1 \times 10^{-4}$, following a warm-up phase covering the initial 5\% of training steps. To accommodate high-resolution ultrasound scans alongside detailed clinical reports, we set a per-device batch size of 1 and extend the maximum sequence length to 16,384 tokens.
 
This SFT paradigm effectively enables UMind-VL to inherit the robust vision-language reasoning capabilities of Qwen3-VL while acquiring deep domain expertise in ultrasound interpretation.

\subsection{Segmentation}
To validate the segmentation capabilities of our multimodal large model, UMind-VL, we conducted a comprehensive quantitative comparison on a diverse benchmark covering 11 distinct anatomical sites in ultrasound imaging. All test data were sourced exclusively from publicly available datasets, including prenatal~\citep{SAPPIA2025103640,correggio2023fetal,jieyun_2024_10829116,ashkani2022fast,van2018automated}, lung~\citep{mclaughlan2024lung}, gynecological~\citep{DBLP:journals/corr/abs-2207-06799}, abdominal~\citep{dataset,xu_yiming_2022_7272660,orlando2020ussimandsegm}, musculoskeletal~\citep{marzola2021transverse}, thyroid~\citep{pedraza2015open,gong2021multi,thyroid_cine_2021}, neurosonography~\citep{ding2022mallesnet,montoya2016ultrasound,tyagi2024nerve,kronke2022tracked}, prostate~\citep{shao2024micro}, breast~\citep{guo2021segmentation,ardakani2023open,hamyoon2022artificial,gomez2024bus,yap2017automated,vallez2024busuclm,iqbal2023busuc,al2020dataset,pawlowska2024curated}, gastrointestinal~\citep{he2022query2} and cardiac~\citep{leclerc2019deep,xu2021cardiacudc,ouyang2020video,reddy2023video}.

We compare UMind-VL against two main categories of baseline models: \textbf{Specialist Models}-these are state-of-the-art models designed specifically for segmentation tasks, including DeepLabV3~\citep{chen2017rethinking}, SegFormer~\citep{xie2021segformer}, UltraSAM~\citep{meyer2025ultrasam}, Mask2Former~\citep{cheng2022masked}; \textbf{Generalist Models}-these are large-scale multimodal models (SA2VA-1B and SA2VA-2B~\citep{yuan2025sa2va}) that can be adapted for segmentation, representing alternative general-purpose approaches. We finetune these generalist models on the union of all public training sets from the aforementioned datasets. The performance is evaluated using Mean Intersection over Union (mIoU, \%).

\begin{table}[htbp]
\centering
\small
\caption{\textbf{Performance comparison on medical segmentation datasets.}}
\label{tab:eval_seg}
\begin{tabular}{@{}lccccccc@{}}
\toprule
\textbf{Anatomical Site} 
& \multicolumn{4}{c}{\textit{Specialist Models}} 
& \multicolumn{2}{c}{\textit{Generalist Models}} 
& \multicolumn{1}{c}{\textit{Our Model}} \\
\cmidrule(r){2-5} \cmidrule(r){6-7} \cmidrule{8-8}
& Deeplabv3 & Segformer & Ultrasam & Mask2former 
& SA2VA(1B) & SA2VA(2B) 
& UMind-VL \\

\midrule
Prenatal         & 85.67 & 80.76 & 78.97 & \textbf{88.13} & 80.26 & 82.75 & 86.81 \\
Lung             & 47.00 & 51.48 & 48.50 & \textbf{57.55} & 46.52 & 44.57 & 52.94 \\
Gynecological    & 67.56 & 47.74 & 64.71 & 69.74 & 69.00 & 71.89 & \textbf{72.38} \\
Abdominal        & 32.63 & 32.01 & 46.03 & 49.46 & 49.23 & 48.17 & \textbf{60.07} \\
Musculoskeletal  & 78.33 & 69.69 & 75.18 & 80.14 & 76.36 & 79.71 & \textbf{81.24} \\
Thyroid          & 65.03 & 52.96 & 69.87 & 71.64 & 61.66 & 65.84 & \textbf{73.50} \\
Neurosonography  & 56.07 & 47.67 & 57.50 & \textbf{71.06} & 50.42 & 59.53 & 65.49 \\
Prostate         & 88.32 & 83.94 & 85.32 & \textbf{89.92} & 85.83 & 86.79 & 89.31 \\
Breast           & 71.33 & 54.15 & 78.43 & 78.53 & 76.51 & 78.00 & \textbf{81.22} \\
Gastrointestinal & 62.24 & 44.75 & 67.22 & 72.15 & 65.19 & 64.95 & \textbf{74.69} \\
Cardiac          & 80.30 & 70.94 & 76.45 & \textbf{81.37} & 75.63 & 75.89 & 80.66 \\
\midrule
Average          & 66.77 & 57.83 & 68.02 & 73.61 & 66.96 & 68.92 & \textbf{74.39} \\
\bottomrule
\end{tabular}
\end{table}

As demonstrated in the table~\ref{tab:eval_seg}, our UMind-VL achieves the highest average mIoU of 74.39\% across all 11 anatomical sites. This result signifies a new state-of-the-art in generalist ultrasound segmentation, confirming our model's superior performance and generalization capabilities.

UMind-VL significantly outperforms other generalist models. It surpasses the SA2VA(2B) model (68.92\% average mIoU) by a substantial margin of 5.47 points and the SA2VA(1B) model (66.96\% average mIoU) by 7.43 points. More notably, UMind-VL also outperforms all specialist segmentation models in average performance. It exceeds the previous best-performing specialist, Mask2former (73.61\%), while demonstrating significantly stronger results than Deeplabv3 (66.77\%), Ultrasam (68.02\%), and Segformer (57.83\%). Analyzing the per-site performance, UMind-VL achieves the highest scores in the majority of categories (6 out of 11).

The quantitative results strongly validate the design of UMind-VL. Its ability to not only perform segmentation—a novel capability for the multimodal models, but also outperform highly optimized specialist models, demonstrates its potential as a powerful and versatile foundational model for multimodal ultrasound image analysis.

\subsection{Detection}
\label{sec:deteval}

To evaluate the detection capability of UMind-VL on medical images, we conduct a comparative analysis against a range of baseline models using the BUS-CoT~\citep{yu2025chain} and TN5k~\citep{zhang2025tn5000} datasets, which are dedicated to breast and thyroid lesion detection, respectively. All test data are drawn from public sources and ensuring no overlap with our training sets. 

The models for comparison are grouped into three categories: Specialist Models, representing state-of-the-art object detectors (Faster-RCNN-R50~\citep{ren2016faster}, Detr-R50~\citep{carion2020end}, Deform-Detr-R50~\citep{zhu2020deformable}), are fine-tuned on the respective training sets of the two datasets; Generalist Models, comprising both leading closed-source (e.g., GPT-5-2025-08-07~\citep{openai2025gpt-5-system-card}, Claude4.5-sonnet-2025-09-29~\citep{anthropic2025claude-sonnet-4-5}, Gemini-2.5-Pro~\citep{comanici2025gemini}) and open-source (e.g., Qwen3-VL-30B-A3B-Instruct~\citep{bai2025qwen3vl}, InternVL3.5-30B-A3B-Instruct~\citep{wang2025internvl3}) multimodal large language models (MLLMs) for general tasks; and Medical Generalist Models, which are MLLMs specialized for the medical domain, including MedVLM-R1-2B~\citep{pan2025medvlm}, Citrus-V-8B-v1.0~\citep{wang2025citrus}, MedGemma-27B~\citep{sellergren2025medgemma}, Lingshu-32B~\citep{xu2025lingshu}, and HuatuoGPT-V-34B~\citep{chen2024huatuogpt}.

While Specialist Models generate both confidence scores and bounding box coordinates as detection outputs, LLM-based models only directly produce bounding box coordinates. In our experiments, we use an Intersection over Union (IoU) threshold of 0.5 to identify true positives. To ensure a consistent evaluation metric across all models, for Specialist Models, we select the point on the Precision-Recall curve that achieves the maximum F1 score to represent the model's performance. The corresponding precision and recall at this point were used to compute the F1 score, referred to as Best F1. For LLM-based models, we directly calculate precision and recall based on model outputs and derive the corresponding F1 score.

\begin{table}
\centering
\small
\caption{\textbf{Performance comparison on open-source medical lesion detection datasets.} }
\label{tab:eval_det}
\begin{tabular}{@{}lcccccc@{}}
\toprule
    \textbf{Models} & \multicolumn{3}{c}{\textbf{BUS-CoT}} & \multicolumn{3}{c}{\textbf{TN5k}} \\
\midrule
    & \textbf{Precision} & \textbf{Recall} & \textbf{(Best) F1} & \textbf{Precision} & \textbf{Recall} & \textbf{(Best) F1} \\
\midrule
\multicolumn{7}{c}{\textit{Specialist Models}} \\
\midrule
Faster-RCNN-R50  & 87.25 & 81.07 & 84.05 & 83.69 & 86.20 & 84.93 \\
Detr-R50  & 92.32 & 86.73 & 89.44 & \textbf{93.76} & 90.46 & 92.08 \\
Deform-Detr-R50  & \textbf{94.76} & 88.70 & 91.63 & 92.91 & \textbf{91.42} & \textbf{92.16} \\
\midrule
\multicolumn{7}{c}{\textit{Generalist Models}} \\
\midrule
GPT-5-2025-08-07 & 31.03 & 29.66 & 30.33 & 10.26 & 10.60 & 10.43 \\
Claude4.5-sonnet-2025-09-29 & 40.19 & 35.64 & 37.78 & 15.52 & 14.60 & 15.04 \\
Gemini-2.5-Pro & 12.27 & 11.74 & 12.00 & 0.80 & 0.80 & 0.80 \\
Qwen3-VL-30B-A3B-Instruct & 65.55 & 59.43 & 62.34 & 32.26 & 31.10 & 31.67 \\
InternVL3.5-30B-A3B-Instruct & 1.66 & 0.52 & 0.80 & 4.25 & 1.70 & 2.43 \\
\midrule
\multicolumn{7}{c}{\textit{Medical Generalist Models}} \\
\midrule
MedVLM-R1-2B  & 13.16 & 0.52 & 1.01 & 3.23 & 0.10 & 0.19 \\
Citrus-V-8B-v1.0  & 5.72 & 5.14 & 5.41 & 2.43 & 2.30 & 2.36 \\
MedGemma-27B  & 9.31 & 8.28 & 8.76 & 3.93 & 3.80 & 3.86 \\
Lingshu-32B  & 14.43 & 10.59 & 12.21 & 5.46 & 3.50 & 4.27 \\
HuatuoGPT-V-34B  & 24.66 & 23.06 & 23.84 & 8.09 & 7.90 & 7.99 \\
\midrule
\multicolumn{7}{c}{\textit{Our Model}} \\
\midrule
UMind-VL & 94.28 & \textbf{89.83} & \textbf{92.00} & 90.92 & 91.10  & 91.01 \\
\bottomrule
\end{tabular}
\end{table}

As evidenced by the results in Table~\ref{tab:eval_det}, a clear performance dichotomy exists among the baselines. Previous Generalist Models and Medical Generalist Models yield unsatisfactory metrics, highlighting their limitations in precise localization tasks. In contrast, Specialist Models exhibit significantly superior performance, thereby remaining the preferred choice for practical deployment in clinical settings. However, UMind-VL bridges this gap, demonstrating detection capabilities highly competitive with state-of-the-art specialist detectors while substantially surpassing all other LLM-based models. On the BUS-CoT dataset, our model achieves an F1 score of 92.00\%, outperforming both the state-of-the-art specialist model, Deform-Detr-R50 (91.63\%), and the most advanced LLM-based counterpart, Qwen3-VL-30B-A3B-Instruct (62.34\%). Notably, it surpasses the latter by a substantial margin of 29.66 points. A similar trend is observed on the TN5k dataset, where UMind-VL achieves an F1 score of 91.01\%, surpassing Faster-RCNN-R50 (84.93\%) and dramatically leading the best LLM-based model, Qwen3-VL-30B-A3B-Instruct (31.67\%), by 59.34 points, albeit slightly behind the other two specialist models, Detr-R50 (92.08\%) and Deform-Detr-R50 (92.16\%).

These experimental results confirm the powerful detection capability of UMind-VL on medical images. Its ability to deliver performance comparable to specialized detection models across different anatomical domains, while significantly outperforming other generalist and medical MLLMs, provides a solid empirical foundation for its application in medical image object detection tasks.

\subsection{Keypoint Localization}
We evaluate the fine-grained localization capabilities of UMind-VL on a challenging set of keypoint detection tasks. This experiment is crucial as it not only tests the model's ability to accurately identify anatomical landmarks, which differs from region-based segmentation or detection, but also its potential to assist in clinical measurements.

We benchmark UMind-VL against leading specialist and generalist models on five diverse ultrasound datasets: Cardiac, Liver, Kidney, Gallbladder and Spectrum. Owing to the scarcity of publicly available ultrasound keypoint datasets, all keypoint-related evaluations are conducted on in-house datasets. 
For the Cardiac, Liver, Kidney, and Gallbladder datasets, which consist of 2D ultrasound images, annotations include both measurement lines and keypoints corresponding to key anatomical structures (e.g., left ventricle anteroposterior diameter, kidney superior-inferior diameter, superior gallbladder pole).
The Spectrum dataset is comprised of cardiac Doppler ultrasound images, where annotations focus on specific peak locations within the spectral waveform.

For specialist models, we use three high-performing models in the field of Pose Estimation, including HRNet~\citep{sun2019deep}, SimCC~\citep{DBLP:conf/eccv/LiYLZWWYX22} and ViTPose~\citep{NEURIPS2022_fbb10d31}. 
For generalist models, we compare against other large multimodal models, including Qwen3-VL~\citep{bai2025qwen3vl}, InternVL3.5~\citep{wang2025internvl3}, and LingShu~\citep{xu2025lingshu}, to assess the performance of general-purpose models on this specialized task. Since these generalist models are not inherently designed for keypoint localization, they are fine-tuned on the five datasets before evaluation. 

Performance is evaluated using a standard keypoint localization metric: Mean Distance Error (MDE).
The results from Table~\ref{tab:eval_point} demonstrate the superiority of UMind-VL in the keypoint localization task. Our model achieves the top result across all five datasets with a lowest average MDE of 14.33, surpassing both highly specialized models and other large-scale generalist models. Notably, despite the high structural similarity with Qwen3-VL-4B, UMind-VL's performance exhibits clear superiority. 
We hypothesize that two factors drive this improvement. First, the diverse training data implicitly expose the model to meaningful anatomical patterns and clinically relevant cues. Second, the proposed light Dynamic Convolutional Mask Decoder encourages the encoder to learn intrinsic, geometry-aware representations. In contrast, segmentation-oriented MLLMs such as SA2VA~\citep{yuan2025sa2va} rely on external expert models for segmentation, which may limit their ability to learn such inherent structural understanding.
These findings, combined with the segmentation and detection results, validate our model as a truly versatile and powerful foundational tool, capable of excelling at both region-level and landmark-level understanding in complex medical imaging tasks.

\begin{table}[tbp]
\centering
\small
\caption{\textbf{Performance comparison on keypoint localization datasets.} }
\label{tab:eval_point}
\begin{tabular}{@{}lcccccc@{}}
\toprule
    & \textbf{Cardiac} & \textbf{Liver} & \textbf{Kidney} & \textbf{Gallbladder} & \textbf{Spectrum} & \textbf{Average} \\
\midrule
\multicolumn{7}{c}{\textit{Specialist Models}} \\
\midrule
HRNet-W32  & 14.15 & 23.64 & 21.34 & 20.83 & 18.30 & 19.65 \\
SimCC-R50  & 18.23 & 36.74 & 45.85 & 38.36 & 18.42 & 31.52\\
ViTPose-S  & 12.57 & 22.91 & 21.94 & 17.85 & 18.57 & 18.77 \\
\midrule
\multicolumn{7}{c}{\textit{Generalist Models}} \\
\midrule
Qwen3-VL-4B-Instruct & 11.85 & 19.18 & 12.32 & \textbf{15.78} & 21.09 & 16.04\\ 
InternVL3.5-2B-Instruct & 11.90 & 19.00 & 19.37 & 19.54 & 18.63 & 17.69\\
Lingshu-7B & 17.05 & 19.16 & 32.27 & 16.72 & 23.03 & 21.65\\
\midrule
\multicolumn{7}{c}{\textit{Our Model}} \\
\midrule
UMind-VL & \textbf{11.44} & \textbf{16.27} & \textbf{10.66} & 16.29 & \textbf{17.00} & \textbf{14.33} \\
\bottomrule
\end{tabular}
\end{table}

\subsection{Diagnosis}

To evaluate lesion diagnosis performance, we test on three datasets: BUS-CoT~\citep{yu2025chain} (the version predating November 1st, 2025), TN5K~\citep{zhang2025tn5000}, and BUS-BRA~\citep{gomez2024bus}. BUS-CoT (breast) and TN5K (thyroid) serve as in-distribution datasets, while BUS-BRA (breast) is used for out-of-distribution (OOD) evaluation. The metric is the classification accuracy of benign versus malignant lesions.

\begin{table}[htpb]
\centering
\small
\caption{\textbf{Performance comparison on diagnosis datasets.} }
\label{tab:eval_diagnosis}
\begin{tabular}{@{}lccc@{}}
\toprule
\textbf{Models} & \textbf{BUS-CoT} & \textbf{TN5K} & \textbf{BUS-BRA} (OOD) \\
\midrule
\multicolumn{4}{c}{\textit{Specialist Models}} \\
\midrule
ResNet50 & 74.52 & 88.80 & \\
Swin-b & 77.42 & \textbf{89.80} & -- \\
Vit-l-16 & 75.05 & 88.40 & \\
\midrule
\multicolumn{4}{c}{\textit{Generalist Models}} \\
\midrule
GPT-5-2025-08-07  & 65.27 & 65.30 & 68.05 \\
Claude4.5-sonnet-2025-09-29  & 59.68 & 50.00 & 25.71 \\
Gemini-2.5-Pro & 57.53 & 71.30 & 46.29 \\
Qwen3-VL-30B-A3B-Instruct & 46.34 & 73.10 & 32.48 \\
InternVL3.5-30B-A3B-Instruct & 59.25 & 46.30 & 59.63 \\
\midrule
\multicolumn{4}{c}{\textit{Medical Generalist Models}} \\
\midrule
MedVLM-R1-2B & 55.05 & 30.90 & 58.40\\
Citrus-V-8B-v1.0 & 46.67 & 74.70 & 32.59 \\
MedGemma-27B & 47.10 & 46.90 & 40.48\\
Lingshu-32B & 55.59 & 69.20 & 34.99 \\
HuatuoGPT-V-34B & 44.41 & 62.60 & 34.51\\
\midrule
\multicolumn{4}{c}{\textit{Our Model}} \\
\midrule 
UMind-VL & \textbf{77.74} & 89.40 & \textbf{84.96}\\
\bottomrule
\end{tabular}
\end{table}

Similar to the detection task, we categorize comparison models into three groups. First, for Specialist Models, we fine-tune ResNet50~\citep{he2016deep}, Swin-B~\citep{liu2021swin}, and ViT-L-16~\citep{dosovitskiy2020image} on BUS-CoT and TN5K data. Second, Generalist and Medical Generalist Models remain the same as in Section \ref{sec:deteval}. Lesion ROIs are provided for all inputs. For generalist models, we employ the prompt: "Determine if the lesion is benign or malignant. Respond exclusively with 'benign' or 'malignant', avoiding any uncertain outcomes." (Translated from Chinese). We use Qwen3-30B-A3B-Instruct-2507~\citep{yang2025qwen3} to post-process responses containing CoT reasoning. Any output indicating uncertainty is considered an error.

Table~\ref{tab:eval_diagnosis} presents the results, supporting three conclusions. First, UMind-VL achieves performance comparable to Specialist Models on in-distribution datasets, even slightly surpassing Swin-B on BUS-CoT (77.74\% vs. 77.42\%) and matching it on TN5K (89.40\% vs. 89.80\%). Second, UMind-VL demonstrates superior generalization on the OOD dataset BUS-BRA (84.96\%), significantly outperforming other models. This suggests that training on large-scale ultrasound data yields better robustness than general pre-training. Third, UMind-VL is robust across varying distributions. While some models exhibit severe class bias (e.g., Qwen3-VL-30B-A3B-Instruct predicts mostly malignant), UMind-VL maintains consistent high accuracy across all datasets regardless of their specific label distributions.

\subsection{Qualitative illustration}

Beyond the quantitative evaluation conducted across the four tasks, we further provide extensive qualitative visualizations of the model’s behavior in Section~\ref{sec:showcase}. These examples cover both single-turn and multi-turn interactions. Across diverse tasks—particularly those involving Grounded Perception—the model demonstrates consistently strong and reliable performance.

\section{Conclusion and Future Work}

In this work, we introduced UMind-VL, a novel ultrasound foundation model that unifies both Grounded Perception and Comprehensive Interpretation within a single framework. To support large-scale multimodal training, we also constructed UMind-DS, a diverse ultrasound dataset comprising 1.2 million image–text pairs. UMind-VL demonstrates strong and consistent performance across a broad spectrum of ultrasound tasks, highlighting its potential as a general-purpose solution for ultrasound intelligence.
In future work, we plan to further enhance the capabilities of UMind-VL, improve the usability and interpretability of its outputs, and advance its translation toward real-world clinical applications.

\clearpage

\setcounter{section}{0} 
\renewcommand{\thesection}{S}
\setcounter{figure}{0}
\captionsetup[figure]{name=Showcase} 
\renewcommand{\thefigure}{\arabic{figure}}

\section{Model Demonstrations}
\label{sec:showcase}

\foreach \name in {
  02seg-1,
  02seg-2,
  03mseg-1,
  03mseg-2,
  04seg,
  05lesion,
  06lesion,
  07lesion,
  08abd,
  09obs,
  10cardiac-1,
  10cardiac-2,
  10cardiac-3,
  11fs-1,
  11fs-2,
  12oodseg-1,
  12oodseg-2}{
    \begin{figure}[h]
        \centering
        \includegraphics[width=0.8\linewidth]{showcases/\name.pdf}

        \IfEndWith{\name}{-2}{
            \ContinuedFloat
            \caption[]{ (Continued)}
        }{
            \IfEndWith{\name}{-3}{
                \ContinuedFloat
                \caption[]{ (Continued)}
            }{
                \caption{} 
            }
        }
        
    \end{figure}
}

\clearpage
\bibliography{biblio}

@article{luo2022biogpt,
  title={BioGPT: generative pre-trained transformer for biomedical text generation and mining},
  author={Luo, Renqian and Sun, Liai and Xia, Yingce and Qin, Tao and Zhang, Sheng and Poon, Hoifung and Liu, Tie-Yan},
  journal={Briefings in bioinformatics},
  volume={23},
  number={6},
  pages={bbac409},
  year={2022},
  publisher={Oxford University Press}
}

@article{singhal2023large,
  title={Large language models encode clinical knowledge},
  author={Singhal, Karan and Azizi, Shekoofeh and Tu, Tao and Mahdavi, S Sara and Wei, Jason and Chung, Hyung Won and Scales, Nathan and Tanwani, Ajay and Cole-Lewis, Heather and Pfohl, Stephen and others},
  journal={Nature},
  volume={620},
  number={7972},
  pages={172--180},
  year={2023},
  publisher={Nature Publishing Group}
}

@article{singhal2025toward,
  title={Toward expert-level medical question answering with large language models},
  author={Singhal, Karan and Tu, Tao and Gottweis, Juraj and Sayres, Rory and Wulczyn, Ellery and Amin, Mohamed and Hou, Le and Clark, Kevin and Pfohl, Stephen R and Cole-Lewis, Heather and others},
  journal={Nature Medicine},
  volume={31},
  number={3},
  pages={943--950},
  year={2025},
  publisher={Nature Publishing Group US New York}
}

@article{dou2025baichuan,
  title={Baichuan-m2: Scaling medical capability with large verifier system},
  author={Dou, Chengfeng and Liu, Chong and Yang, Fan and Li, Fei and Jia, Jiyuan and Chen, Mingyang and Ju, Qiang and Wang, Shuai and Dang, Shunya and Li, Tianpeng and others},
  journal={arXiv preprint arXiv:2509.02208},
  year={2025}
}

@article{sellergren2025medgemma,
  title={Medgemma technical report},
  author={Sellergren, Andrew and Kazemzadeh, Sahar and Jaroensri, Tiam and Kiraly, Atilla and Traverse, Madeleine and Kohlberger, Timo and Xu, Shawn and Jamil, Fayaz and Hughes, C{\'\i}an and Lau, Charles and others},
  journal={arXiv preprint arXiv:2507.05201},
  year={2025}
}

@article{xu2025lingshu,
  title={Lingshu: A Generalist Foundation Model for Unified Multimodal Medical Understanding and Reasoning},
  author={Xu, Weiwen and Chan, Hou Pong and Li, Long and Aljunied, Mahani and Yuan, Ruifeng and Wang, Jianyu and Xiao, Chenghao and Chen, Guizhen and Liu, Chaoqun and Li, Zhaodonghui and others},
  journal={arXiv preprint arXiv:2506.07044},
  year={2025}
}

@article{tang2025ufo,
  title={Ufo: A unified approach to fine-grained visual perception via open-ended language interface},
  author={Tang, Hao and Xie, Chenwei and Wang, Haiyang and Bao, Xiaoyi and Weng, Tingyu and Li, Pandeng and Zheng, Yun and Wang, Liwei},
  journal={arXiv preprint arXiv:2503.01342},
  year={2025}
}

@article{marcinkevivcs2023regensburg,
  title={Regensburg pediatric appendicitis dataset},
  author={Marcinkevi{\v{c}}s, Ri{\v{c}}ards and Reis Wolfertstetter, Patricia and Klimiene, Ugne and Chin-Cheong, Kieran and Paschke, Alyssia and Zerres, Julia and Denzinger, Markus and Niederberger, David and Wellmann, Sven and Ozkan, Ece and others},
  journal={(No Title)},
  year={2023},
  publisher={Zenodo}
}

@article{byra2018transfer,
  title={Transfer learning with deep convolutional neural network for liver steatosis assessment in ultrasound images},
  author={Byra, Micha{\l} and Styczynski, Grzegorz and Szmigielski, Cezary and Kalinowski, Piotr and Micha{\l}owski, {\L}ukasz and Paluszkiewicz, Rafa{\l} and Ziarkiewicz-Wr{\'o}blewska, Bogna and Zieniewicz, Krzysztof and Sobieraj, Piotr and Nowicki, Andrzej},
  journal={International journal of computer assisted radiology and surgery},
  volume={13},
  number={12},
  pages={1895--1903},
  year={2018},
  publisher={Springer}
}

@inproceedings{zhang2022busis,
  title={BUSIS: a benchmark for breast ultrasound image segmentation},
  author={Zhang, Yingtao and Xian, Min and Cheng, Heng-Da and Shareef, Bryar and Ding, Jianrui and Xu, Fei and Huang, Kuan and Zhang, Boyu and Ning, Chunping and Wang, Ying},
  booktitle={Healthcare},
  volume={10},
  number={4},
  pages={729},
  year={2022},
  organization={MDPI}
}

@inproceedings{AUITD,
  title={Algerian Ultrasound Images Thyroid Dataset: AUITD},
  author={AZOUZ Maroua},
  booktitle={https://www.kaggle.com/datasets/azouzmaroua/algeria-ultrasound-images-thyroid-dataset-auitd},
  volume={},
  number={},
  pages={},
  year={2022},
  organization={Kaggle}
}

@inproceedings{wang2022key,
  title={Key-frame guided network for thyroid nodule recognition using ultrasound videos},
  author={Wang, Yuchen and Li, Zhongyu and Cui, Xiangxiang and Zhang, Liangliang and Luo, Xiang and Yang, Meng and Chang, Shi},
  booktitle={International Conference on Medical Image Computing and Computer-Assisted Intervention},
  pages={238--247},
  year={2022},
  organization={Springer}
}

@InProceedings{10.1007/978-3-031-16437-8_59,
author="Lin, Zhi
and Lin, Junhao
and Zhu, Lei
and Fu, Huazhu
and Qin, Jing
and Wang, Liansheng",
editor="Wang, Linwei
and Dou, Qi
and Fletcher, P. Thomas
and Speidel, Stefanie
and Li, Shuo",
title="A New Dataset and a Baseline Model for Breast Lesion Detection in Ultrasound Videos",
booktitle="Medical Image Computing and Computer Assisted Intervention -- MICCAI 2022",
year="2022",
publisher="Springer Nature Switzerland",
address="Cham",
pages="614--623",
}

@ARTICLE{10015121,
  author={Mo, Yuhao and Han, Chu and Liu, Yu and Liu, Min and Shi, Zhenwei and Lin, Jiatai and Zhao, Bingchao and Huang, Chunwang and Qiu, Bingjiang and Cui, Yanfen and Wu, Lei and Pan, Xipeng and Xu, Zeyan and Huang, Xiaomei and Li, Zhenhui and Liu, Zaiyi and Wang, Ying and Liang, Changhong},
  journal={IEEE Transactions on Medical Imaging}, 
  title={HoVer-Trans: Anatomy-aware HoVer-Transformer for ROI-free Breast Cancer Diagnosis in Ultrasound Images}, 
  year={2023},
  volume={},
  number={},
  pages={1-1},
  doi={10.1109/TMI.2023.3236011}}

@inproceedings{leenings2024c,
  title={C-trus: A novel dataset and initial benchmark for colon wall segmentation in transabdominal ultrasound},
  author={Leenings, Ramona and Konowski, Maximilian and Winter, Nils R and Ernsting, Jan and Fisch, Lukas and Barkhau, Carlotta and Dannlowski, Udo and L{\"u}gering, Andreas and Jiang, Xiaoyi and Hahn, Tim},
  booktitle={International Workshop on Advances in Simplifying Medical Ultrasound},
  pages={101--111},
  year={2024},
  organization={Springer}
}

@article{wiedemann2025covid,
  author={Wiedemann, Nina and Boer, Dianne de Korte-de and Richter, Matthias and van de Weijer, Sjors and Buhre, Charlotte and Eggert, Franz A. M. and Aarnoudse, Sophie and Grevendonk, Lotte and Röber, Steffen and Remie, Carlijn M.E. and Buhre, Wolfgang and Henry, Ronald and Born, Jannis},
  journal={IEEE Journal of Biomedical and Health Informatics}, 
  title={COVID-BLUeS - A Prospective Study on the Value of AI in Lung Ultrasound Analysis}, 
  year={2025},
  volume={29},
  number={9},
  pages={6301-6310},
  doi={10.1109/JBHI.2025.3543686}
}

@article{divekar2024leveraging,
  title={Leveraging ai for automatic classification of pcos using ultrasound imaging},
  author={Divekar, Atharva and Sonawane, Atharva},
  journal={arXiv preprint arXiv:2501.01984},
  year={2024}
}

@article{sendra2023maternal,
  title={Maternal fetal ultrasound planes from low-resource imaging settings in five African countries},
  author={Sendra-Balcells, C and Campello, VM and Torrents-Barrena, J and Ahmed, YA and Elattar, M and Ohene-Botwe, B and Nyangulu, P and Stones, W and Ammar, M and Benamer, LN and others},
  journal={Zenodo},
  year={2023}
}

@article{prabakaran2023fpus23,
  title={FPUS23: an ultrasound fetus phantom dataset with deep neural network evaluations for fetus orientations, fetal planes, and anatomical features},
  author={Prabakaran, Bharath Srinivas and Hamelmann, Paul and Ostrowski, Erik and Shafique, Muhammad},
  journal={IEEE Access},
  volume={11},
  pages={58308--58317},
  year={2023},
  publisher={IEEE}
}

@misc{burgosfetal_planes_db,
  title={FETAL\_PLANES\_DB: Common maternal-fetal ultrasound images (Jun 2020)},
  author={Burgos-Artizzu, XP and Coronado-Gutierrez, D and Valenzuela-Alcaraz, B and Bonet-Carne, E and Eixarch, E and Crispi, F and Gratac{\'o}s, E}
}

@inproceedings{bi2023mi,
  title={MI-SegNet: Mutual information-based US segmentation for unseen domain generalization},
  author={Bi, Yuan and Jiang, Zhongliang and Clarenbach, Ricarda and Ghotbi, Reza and Karlas, Angelos and Navab, Nassir},
  booktitle={International Conference on Medical Image Computing and Computer-Assisted Intervention},
  pages={130--140},
  year={2023},
  organization={Springer}
}

@article{meiburger2021carotid,
  title={Carotid ultrasound boundary study (CUBS): An open multicenter analysis of computerized intima--media thickness measurement systems and their clinical impact},
  author={Meiburger, Kristen M and Zahnd, Guillaume and Faita, Francesco and Loizou, Christos P and Carvalho, Catarina and Steinman, David A and Gibello, Lorenzo and Bruno, Rosa Maria and Marzola, Francesco and Clarenbach, Ricarda and others},
  journal={Ultrasound in Medicine \& Biology},
  volume={47},
  number={8},
  pages={2442--2455},
  year={2021},
  publisher={Elsevier}
}

@article{li2023dsmt,
  title={Dsmt-net: dual self-supervised multi-operator transformation for multi-source endoscopic ultrasound diagnosis},
  author={Li, Jiajia and Zhang, Pingping and Wang, Teng and Zhu, Lei and Liu, Ruhan and Yang, Xia and Wang, Kaixuan and Shen, Dinggang and Sheng, Bin},
  journal={IEEE Transactions on Medical Imaging},
  volume={43},
  number={1},
  pages={64--75},
  year={2023},
  publisher={IEEE}
}

@article{tseng2024real,
  title={Real-time automatic m-mode echocardiography measurement with panel attention},
  author={Tseng, Ching-Hsun and Chien, Shao-Ju and Wang, Po-Shen and Lee, Shin-Jye and Pu, Bin and Zeng, Xiao-Jun},
  journal={IEEE Journal of Biomedical and Health Informatics},
  volume={28},
  number={9},
  pages={5383--5395},
  year={2024},
  publisher={IEEE}
}

@article{katumba2025dataset,
  title={A dataset of lung ultrasound images for automated AI-based lung disease classification},
  author={Katumba, Andrew and Murindanyi, Sudi and Okila, Nixson and Nakatumba-Nabende, Joyce and Mwikirize, Cosmas and Serugunda, Jonathan and Bugeza, Samuel and Oriekot, Anthony and Bossa, Juliet and Nabawanuka, Eva},
  journal={Data in Brief},
  pages={112034},
  year={2025},
  publisher={Elsevier}
}

@article{uterine-fibroid-ultrasound,
  title={uterine fibroid ultrasound images},
  author={Tiantian Yang},
  journal={Mendeley Data, V2, doi: 10.17632/n2zcmcypgb.2},
  year={2023}
}

@article{FUGC,
  title={A Dataset for Fetal Ultrasound Grand Challenge: Semi-Supervised Cervical Segmentation},
  author={Jieyun Bai},
  journal={Zenodo, https://doi.org/10.5281/zenodo.16893174.},
  year={2025}
}

@article{gonzalez2024methods,
  title={Methods for the acquisition, learning-based segmentation and quantitative analysis of ultrasound volumes},
  author={Gonzalez Duque, Vanessa},
  year={2024},
  school={Technische Universit{\"a}t M{\"u}nchen}
}

@article{USAnotAI,
  title={USAnotAI: Organ Classification on Abdominal Ultrasound using Javascript},
  author={Kim Ann},
  journal={https://github.com/ftsvd/USAnotAI.},
  year={2019},
}

@inproceedings{carnahan2023segmentation,
  title={Segmentation of the mitral valve from 3D transesophageal echocardiography},
  author={Carnahan, P and Bharucha, A and Eskandari, M and Chen, E and Peters, T},
  booktitle={International Conference On Medical Image Computing And Computer Assisted Intervention (MICCAI)},
  year={2023}
}

@article{he2022query2,
  title={Query2: Query over queries for improving gastrointestinal stromal tumour detection in an endoscopic ultrasound},
  author={He, Qi and Bano, Sophia and Liu, Jing and Liu, Wentian and Stoyanov, Danail and Zuo, Siyang},
  journal={Computers in Biology and Medicine},
  pages={106424},
  year={2022},
  publisher={Elsevier}
}

@article{shao2024micro,
  title={Micro-ultrasound prostate segmentation dataset},
  author={Shao, W and Brisbane, W},
  journal={URL: https://doi. org/10.5281/zenodo},
  volume={10475293},
  year={2024}
}

@article{mclaughlan2024lung,
  title={Lung ultrasound covid phantom dataset used for training machine learning model},
  author={McLaughlan, James R and Howell, Lewis and Ingram, Nicola},
  year={2024},
  publisher={University of Leeds}
}

@article{pawlowska2024curated,
  title={Curated benchmark dataset for ultrasound based breast lesion analysis},
  author={Paw{\l}owska, Anna and {\'C}wierz-Pie{\'n}kowska, Anna and Domalik, Agnieszka and Jagu{\'s}, Dominika and Kasprzak, Piotr and Matkowski, Rafa{\l} and Fura, {\L}ukasz and Nowicki, Andrzej and {\.Z}o{\l}ek, Norbert},
  journal={Scientific Data},
  volume={11},
  number={1},
  pages={148},
  year={2024},
  publisher={Nature Publishing Group UK London}
}

@article{ardakani2023open,
  title={An open-access breast lesion ultrasound image database: Applicable in artificial intelligence studies},
  author={Ardakani, Ali Abbasian and Mohammadi, Afshin and Mirza-Aghazadeh-Attari, Mohammad and Acharya, U Rajendra},
  journal={Computers in Biology and Medicine},
  volume={152},
  pages={106438},
  year={2023},
  publisher={Elsevier}
}

@article{hamyoon2022artificial,
  title={Artificial intelligence, BI-RADS evaluation and morphometry: A novel combination to diagnose breast cancer using ultrasonography, results from multi-center cohorts},
  author={Hamyoon, Hessam and Chan, Wai Yee and Mohammadi, Afshin and Kuzan, Taha Yusuf and Mirza-Aghazadeh-Attari, Mohammad and Leong, Wai Ling and Altintoprak, K{\"u}bra Murzoglu and Vijayananthan, Anushya and Rahmat, Kartini and Ab Mumin, Nazimah and others},
  journal={European Journal of Radiology},
  volume={157},
  pages={110591},
  year={2022},
  publisher={Elsevier}
}

@article{homayoun2022applications,
  title={Applications of machine-learning algorithms for prediction of benign and malignant breast lesions using ultrasound radiomics signatures: A multi-center study},
  author={Homayoun, Hassan and Chan, Wai Yee and Kuzan, Taha Yusuf and Leong, Wai Ling and Altintoprak, K{\"u}bra Murzoglu and Mohammadi, Afshin and Vijayananthan, Anushya and Rahmat, Kartini and Leong, Sook Sam and Mirza-Aghazadeh-Attari, Mohammad and others},
  journal={Biocybernetics and Biomedical Engineering},
  volume={42},
  number={3},
  pages={921--933},
  year={2022},
  publisher={Elsevier}
}

@misc{iqbal2023busuc,
  author = {Iqbal, Ahmed},
  title = {BUS\_UC},
  year = {2023},
  publisher = {Mendeley Data},
  version = {V1},
  doi = {10.17632/3ksd7w7jkx.1}
}

@misc{vallez2024busuclm,
  author       = {Vallez, Noelia and Bueno, Gloria and Deniz, Oscar and Rienda, Miguel Angel and Pastor, Carlos},
  title        = {{BUS-UCLM}: Breast ultrasound lesion segmentation dataset},
  year         = {2024},
  publisher    = {Mendeley Data},
  version      = {V1},
  doi          = {10.17632/7fvgj4jsp7.1}
}

@article{gomez2024bus,
  title={BUS-BRA: a breast ultrasound dataset for assessing computer-aided diagnosis systems},
  author={G{\'o}mez-Flores, Wilfrido and Gregorio-Calas, Maria Julia and Coelho de Albuquerque Pereira, Wagner},
  journal={Medical Physics},
  volume={51},
  number={4},
  pages={3110--3123},
  year={2024},
  publisher={Wiley Online Library}
}

@article{yap2017automated,
  title={Automated breast ultrasound lesions detection using convolutional neural networks},
  author={Yap, Moi Hoon and Pons, Gerard and Marti, Joan and Ganau, Sergi and Sentis, Melcior and Zwiggelaar, Reyer and Davison, Adrian K and Marti, Robert},
  journal={IEEE journal of biomedical and health informatics},
  volume={22},
  number={4},
  pages={1218--1226},
  year={2017},
  publisher={IEEE}
}

@article{al2020dataset,
  title={Dataset of breast ultrasound images},
  author={Al-Dhabyani, Walid and Gomaa, Mohammed and Khaled, Hussien and Fahmy, Aly},
  journal={Data in brief},
  volume={28},
  pages={104863},
  year={2020},
  publisher={Elsevier}
}

@article{guo2021segmentation,
  title={Segmentation and recognition of breast ultrasound images based on an expanded U-Net},
  author={Guo, Yanjun and Duan, Xingguang and Wang, Chengyi and Guo, Huiqin},
  journal={Plos one},
  volume={16},
  number={6},
  pages={e0253202},
  year={2021},
  publisher={Public Library of Science San Francisco, CA USA}
}

@misc{stuhospital2020,
  title        = {STU-Hospital Dataset},
  author = {xbhlk},
  howpublished = {https://github.com/xbhlk/STU-Hospital},
  note         = {Accessed: 2020-02-06}
}

@inproceedings{pedraza2015open,
  title={An open access thyroid ultrasound image database},
  author={Pedraza, Lina and Vargas, Carlos and Narv{\'a}ez, Fabi{\'a}n and Dur{\'a}n, Oscar and Mu{\~n}oz, Emma and Romero, Eduardo},
  booktitle={10th International symposium on medical information processing and analysis},
  volume={9287},
  pages={188--193},
  year={2015},
  organization={SPIE}
}

@article{kronke2022tracked,
  title={Tracked 3D ultrasound and deep neural network-based thyroid segmentation reduce interobserver variability in thyroid volumetry},
  author={Kr{\"o}nke, Markus and Eilers, Christine and Dimova, Desislava and K{\"o}hler, Melanie and Buschner, Gabriel and Schweiger, Lilit and Konstantinidou, Lemonia and Makowski, Marcus and Nagarajah, James and Navab, Nassir and others},
  journal={Plos one},
  volume={17},
  number={7},
  pages={e0268550},
  year={2022},
  publisher={Public Library of Science San Francisco, CA USA}
}

@inproceedings{gong2021multi,
  title={Multi-task learning for thyroid nodule segmentation with thyroid region prior},
  author={Gong, Haifan and Chen, Guanqi and Wang, Ranran and Xie, Xiang and Mao, Mingzhi and Yu, Yizhou and Chen, Fei and Li, Guanbin},
  booktitle={2021 IEEE 18th international symposium on biomedical imaging (ISBI)},
  pages={257--261},
  year={2021},
  organization={IEEE}
}

@misc{thyroid_cine_2021,
  title        = {Thyroid Ultrasound Cine-clip},
  howpublished = {\url{https://stanfordaimi.azurewebsites.net/datasets/a72f2b02-7b53-4c5d-963c-d7253220bfd5}},
  note         = {Accessed: 2021-10-09}
}

@article{zhang2025tn5000,
  title={TN5000: An Ultrasound Image Dataset for Thyroid Nodule Detection and Classification},
  author={Zhang, Huan and Liu, Qianglin and Han, Xiaolin and Niu, Lijuan and Sun, Weidong},
  journal={Scientific data},
  volume={12},
  number={1},
  pages={1437},
  year={2025},
  publisher={Nature Publishing Group UK London}
}

@misc{dataset,
    author = {Egger, Jan},
    year = {2018},
    month = {12},
    pages = {},
    title = {100+ 2D US Images and Tumor Segmentation Masks},
    doi = {10.13140/RG.2.2.36586.77761}
}

@misc{orlando2020ussimandsegm,
  author       = {Orlando, José Ignacio and Vitale, Santiago},
  title        = {{US} simulation \& segmentation: real and synthetic abdominal ultrasound scans, with manual segmentations},
  year         = {2020},
  howpublished = {\url{https://www.kaggle.com/datasets/ignaciorlando/ussimandsegm}}
}

@misc{xu_yiming_2022_7272660,
  author       = {Xu Yiming and
                  Zheng Bowen and
                  Liu Xiaohong and
                  Wu Tao and
                  Ju Jinxiu and
                  Wang Shijie and
                  Lian Yufan and
                  Zhang Hongjun and
                  Liang Tong and
                  Sang Ye and
                  Jiang Rui and
                  Wang Guangyu and
                  Ren Jie and
                  Chen Ting},
  title        = {Annotated Ultrasound Liver images},
  month        = nov,
  year         = 2022,
  publisher    = {Zenodo},
  doi          = {10.5281/zenodo.7272660},
  url          = {https://doi.org/10.5281/zenodo.7272660},
}

@inproceedings{singla2023open,
  title={The open kidney ultrasound data set},
  author={Singla, Rohit and Ringstrom, Cailin and Hu, Grace and Lessoway, Victoria and Reid, Janice and Nguan, Christopher and Rohling, Robert},
  booktitle={International Workshop on Advances in Simplifying Medical Ultrasound},
  pages={155--164},
  year={2023},
  organization={Springer}
}

@misc{momot2022cca,
  author = {Momot, Agata},
  title = {Common Carotid Artery Ultrasound Images},
  year = {2022},
  publisher = {Mendeley Data},
  version = {V1},
  doi = {10.17632/d4xt63mgjm.1}
}

@article{leclerc2019deep,
  title={Deep learning for segmentation using an open large-scale dataset in 2D echocardiography},
  author={Leclerc, Sarah and Smistad, Erik and Pedrosa, Joao and {\O}stvik, Andreas and Cervenansky, Frederic and Espinosa, Florian and Espeland, Torvald and Berg, Erik Andreas Rye and Jodoin, Pierre-Marc and Grenier, Thomas and others},
  journal={IEEE transactions on medical imaging},
  volume={38},
  number={9},
  pages={2198--2210},
  year={2019},
  publisher={IEEE}
}

@misc{xu2021cardiacudc,
  author = {Xu, Xiaowei},
  title = {CardiacUDC Dataset: Graph-Driven Unsupervised Domain Adaptation for Echocardiogram Video},
  year = {2021},
  howpublished = {Kaggle},
  url = {https://www.kaggle.com/datasets/xiaoweixumedicalai/cardiacudc-dataset}
}

@article{ouyang2020video,
  title={Video-based AI for beat-to-beat assessment of cardiac function},
  author={Ouyang, David and He, Bryan and Ghorbani, Amirata and Yuan, Neal and Ebinger, Joseph and Langlotz, Curtis P and Heidenreich, Paul A and Harrington, Robert A and Liang, David H and Ashley, Euan A and others},
  journal={Nature},
  volume={580},
  number={7802},
  pages={252--256},
  year={2020},
  publisher={Nature Publishing Group}
}

@article{reddy2023video,
  title={Video-based deep learning for automated assessment of left ventricular ejection fraction in pediatric patients},
  author={Reddy, Charitha D and Lopez, Leo and Ouyang, David and Zou, James Y and He, Bryan},
  journal={Journal of the American Society of Echocardiography},
  volume={36},
  number={5},
  pages={482--489},
  year={2023},
  publisher={Elsevier}
}

@article{duffy2022high,
  title={High-throughput precision phenotyping of left ventricular hypertrophy with cardiovascular deep learning},
  author={Duffy, Grant and Cheng, Paul P and Yuan, Neal and He, Bryan and Kwan, Alan C and Shun-Shin, Matthew J and Alexander, Kevin M and Ebinger, Joseph and Lungren, Matthew P and Rader, Florian and others},
  journal={JAMA cardiology},
  volume={7},
  number={4},
  pages={386--395},
  year={2022},
  publisher={American Medical Association}
}

@article{huang2022fix,
  title={Fix-a-step: Semi-supervised learning from uncurated unlabeled data},
  author={Huang, Zhe and Sidhom, Mary-Joy and Wessler, Benjamin S and Hughes, Michael C},
  journal={arXiv preprint arXiv:2208.11870},
  year={2022}
}

@misc{unity2022,
  title        = {Unity Imaging Collaborative: Open-access datasets, models, and code for the development and validation of {AI} in cardiology},
  year         = {2022},
  author       = {Shun-Shin, Matthew},
  url          = {https://data.unityimaging.net}
}

@article{SAPPIA2025103640,
title = {ACOUSLIC-AI challenge report: Fetal abdominal circumference measurement on blind-sweep ultrasound data from low-income countries},
journal = {Medical Image Analysis},
volume = {105},
pages = {103640},
year = {2025},
issn = {1361-8415},
doi = {https://doi.org/10.1016/j.media.2025.103640},
url = {https://www.sciencedirect.com/science/article/pii/S1361841525001872},
author = {M. Sofia Sappia and Chris L. {de Korte} and Bram {van Ginneken} and Dean Ninalga and Satoshi Kondo and Satoshi Kasai and Kousuke Hirasawa and Tanya Akumu and Carlos Martín-Isla and Karim Lekadir and Victor M. Campello and Jorge Fabila and Anette Beverdam and Jeroen {van Dillen} and Chase Neff and Keelin Murphy}
}

@misc{correggio2023fetal,
  author = {Da Correggio, Karine Souza and Noya Galluzzo, Roberto and Santos, Luís Otávio and Soares Muylaert Barroso, Felipe and Zimmermann Loureiro Chaves, Thiago and Sherlley Casimiro Onofre, Alexandre and von Wangenheim, Aldo},
  title = {Fetal Abdominal Structures Segmentation Dataset Using Ultrasonic Images},
  year = {2023},
  publisher = {Mendeley Data},
  version = {V1},
  doi = {10.17632/4gcpm9dsc3.1}
}

@article{ashkani2022fast,
  title={Fast and accurate U-net model for fetal ultrasound image segmentation},
  author={Ashkani Chenarlogh, Vahid and Ghelich Oghli, Mostafa and Shabanzadeh, Ali and Sirjani, Nasim and Akhavan, Ardavan and Shiri, Isaac and Arabi, Hossein and Sanei Taheri, Morteza and Tarzamni, Mohammad Kazem},
  journal={Ultrasonic imaging},
  volume={44},
  number={1},
  pages={25--38},
  year={2022},
  publisher={SAGE Publications Sage CA: Los Angeles, CA}
}

@misc{jieyun_2024_10829116,
  author       = {Jieyun, Bai and
                  ZhanHong, Ou},
  title        = {Pubic Symphysis-Fetal Head Segmentation and Angle
                   of Progression
                  },
  month        = mar,
  year         = 2024,
  publisher    = {Zenodo},
  version      = {v2},
  doi          = {10.5281/zenodo.10829116},
  url          = {https://doi.org/10.5281/zenodo.10829116},
}

@article{van2018automated,
  title={Automated measurement of fetal head circumference using 2D ultrasound images},
  author={van den Heuvel, Thomas LA and de Bruijn, Dagmar and de Korte, Chris L and Ginneken, Bram van},
  journal={PloS one},
  volume={13},
  number={8},
  pages={e0200412},
  year={2018},
  publisher={Public Library of Science San Francisco, CA USA}
}

@article{DBLP:journals/corr/abs-2207-06799,
  author    = {Qi Zhao and
               Shuchang Lyu and
               Wenpei Bai and
               Linghan Cai and
               Binghao Liu and
               Meijing Wu and
               Xiubo Sang and
               Min Yang and
               Lijiang Chen},
  title     = {A Multi-Modality Ovarian Tumor Ultrasound Image Dataset for Unsupervised
               Cross-Domain Semantic Segmentation},
  journal   = {CoRR},
  volume    = {abs/2207.06799},
  year      = {2022},
}

@misc{marzola2021transverse,
  author = {Marzola, Francesco and van Alfen, Nens and Doorduin, Jonne and Meiburger, Kristen},
  title = {DATASET for "Deep learning segmentation of transverse musculoskeletal ultrasound images for neuromuscular disease assessment"},
  year = {2021},
  publisher = {Mendeley Data},
  version = {V1},
  doi = {10.17632/3jykz7wz8d.1}
}

@article{cunningham2018estimating,
  title={Estimating full regional skeletal muscle fibre orientation from B-mode ultrasound images using convolutional, residual, and deconvolutional neural networks},
  author={Cunningham, Ryan and S{\'a}nchez, Mar{\'\i}a B and May, Gregory and Loram, Ian},
  journal={Journal of Imaging},
  volume={4},
  number={2},
  pages={29},
  year={2018},
  publisher={MDPI}
}

@article{cronin2020automated,
  title={Automated analysis of musculoskeletal ultrasound images using deep learning},
  author={Cronin, Neil},
  year={2020}
}

@inproceedings{michard2021aw,
  title={AW-Net: Automatic muscle structure analysis on B-mode ultrasound images for injury prevention},
  author={Michard, Hugo and Luvison, Bertrand and Pham, Quoc-Cuong and Morales-Artacho, Antonio J and Guilhem, Ga{\"e}l},
  booktitle={Proceedings of the 12th ACM International Conference on Bioinformatics, Computational Biology, and Health Informatics},
  pages={1--9},
  year={2021}
}

@inproceedings{tyagi2024nerve,
  title={Nerve block target localization and needle guidance for autonomous robotic ultrasound guided regional anesthesia},
  author={Tyagi, Abhishek and Tyagi, Abhay and Kaur, Manpreet and Aggarwal, Richa and Soni, Kapil D and Sivaswamy, Jayanthi and Trikha, Anjan},
  booktitle={2024 IEEE/RSJ International Conference on Intelligent Robots and Systems (IROS)},
  pages={5867--5872},
  year={2024},
  organization={IEEE}
}

@article{ding2022mallesnet,
  title={MallesNet: A multi-object assistance based network for brachial plexus segmentation in ultrasound images},
  author={Ding, Yi and Member, IEEE and Yang, Qiqi and Wang, Yiqian and Chen, Dajiang and Qin, Zhiguang and Zhang, Jian},
  journal={Medical Image Analysis},
  volume={80},
  pages={102511},
  year={2022},
  publisher={Elsevier}
}

@misc{montoya2016ultrasound,
  author       = {Montoya, Anna},
  title        = {Ultrasound Nerve Segmentation: Identify nerve structures in ultrasound images of the neck},
  year         = {2016},
  howpublished = {Kaggle competition dataset},
  url          = {https://www.kaggle.com/c/ultrasound-nerve-segmentation/data}
}

@inproceedings{baum2023mr,
  title={MR to ultrasound registration for prostate challenge-dataset},
  author={Baum, Zachary and Saeed, Shaheer and Min, Zhe and Hu, Yipeng and Barratt, Dean},
  booktitle={Medical Image Computing and Computer Assisted Intervention--MICCAI},
  volume={2023},
  year={2023}
}

@inproceedings{kazemzadeh2014referitgame,
  title={Referitgame: Referring to objects in photographs of natural scenes},
  author={Kazemzadeh, Sahar and Ordonez, Vicente and Matten, Mark and Berg, Tamara},
  booktitle={Proceedings of the 2014 conference on empirical methods in natural language processing (EMNLP)},
  pages={787--798},
  year={2014}
}

@InProceedings{Wang_2024_CVPR,
    author    = {Wang, Junchi and Ke, Lei},
    title     = {LLM-Seg: Bridging Image Segmentation and Large Language Model Reasoning},
    booktitle = {Proceedings of the IEEE/CVF Conference on Computer Vision and Pattern Recognition (CVPR) Workshops},
    month     = {June},
    year      = {2024},
    pages     = {1765-1774}
}

@article{ungi2020automatic,
  title={Automatic spine ultrasound segmentation for scoliosis visualization and measurement},
  author={Ungi, Tamas and Greer, Hastings and Sunderland, Kyle R and Wu, Victoria and Baum, Zachary MC and Schlenger, Christopher and Oetgen, Matthew and Cleary, Kevin and Aylward, Stephen R and Fichtinger, Gabor},
  journal={IEEE Transactions on Biomedical Engineering},
  volume={67},
  number={11},
  pages={3234--3241},
  year={2020},
  publisher={IEEE}
}

@article{yu2025chain,
  title={A Chain-of-thought Reasoning Breast Ultrasound Dataset Covering All Histopathology Categories},
  author={Yu, Haojun and Li, Youcheng and Niu, Zihan and Zhang, Nan and Gong, Xuantong and Li, Huan and Zou, Zhiying and Qi, Haifeng and Cao, Zhenxiao and Lan, Zijie and others},
  journal={arXiv preprint arXiv:2509.17046},
  year={2025}
}

@article{liberman2002breast,
  title={Breast imaging reporting and data system (BI-RADS)},
  author={Liberman, Laura and Menell, Jennifer H},
  journal={Radiologic Clinics},
  volume={40},
  number={3},
  pages={409--430},
  year={2002},
  publisher={Elsevier}
}

@article{grant2015thyroid,
  title={Thyroid ultrasound reporting lexicon: white paper of the ACR thyroid imaging, reporting and data system (TIRADS) committee},
  author={Grant, Edward G and Tessler, Franklin N and Hoang, Jenny K and Langer, Jill E and Beland, Michael D and Berland, Lincoln L and Cronan, John J and Desser, Terry S and Frates, Mary C and Hamper, Ulrike M and others},
  journal={Journal of the American college of radiology},
  volume={12},
  number={12},
  pages={1272--1279},
  year={2015},
  publisher={Elsevier}
}

@inproceedings{chen2024internvl,
  title={Internvl: Scaling up vision foundation models and aligning for generic visual-linguistic tasks},
  author={Chen, Zhe and Wu, Jiannan and Wang, Wenhai and Su, Weijie and Chen, Guo and Xing, Sen and Zhong, Muyan and Zhang, Qinglong and Zhu, Xizhou and Lu, Lewei and others},
  booktitle={Proceedings of the IEEE/CVF conference on computer vision and pattern recognition},
  pages={24185--24198},
  year={2024}
}

@article{bai2025qwen2,
  title={Qwen2. 5-vl technical report},
  author={Bai, Shuai and Chen, Keqin and Liu, Xuejing and Wang, Jialin and Ge, Wenbin and Song, Sibo and Dang, Kai and Wang, Peng and Wang, Shijie and Tang, Jun and others},
  journal={arXiv preprint arXiv:2502.13923},
  year={2025}
}

@misc{openai2025gptoss120bgptoss20bmodel,
      title={gpt-oss-120b \& gpt-oss-20b Model Card}, 
      author={OpenAI},
      year={2025},
      eprint={2508.10925},
      archivePrefix={arXiv},
      primaryClass={cs.CL},
      url={https://arxiv.org/abs/2508.10925}, 
}

@misc{liu2024llavanext,
    title={LLaVA-NeXT: Improved reasoning, OCR, and world knowledge},
    url={https://llava-vl.github.io/blog/2024-01-30-llava-next/},
    author={Liu, Haotian and Li, Chunyuan and Li, Yuheng and Li, Bo and Zhang, Yuanhan and Shen, Sheng and Lee, Yong Jae},
    month={January},
    year={2024}
}

@misc{liu2023improvedllava,
      title={Improved Baselines with Visual Instruction Tuning}, 
      author={Liu, Haotian and Li, Chunyuan and Li, Yuheng and Lee, Yong Jae},
      publisher={arXiv:2310.03744},
      year={2023},
}

@misc{liu2023llava,
      title={Visual Instruction Tuning}, 
      author={Liu, Haotian and Li, Chunyuan and Wu, Qingyang and Lee, Yong Jae},
      publisher={NeurIPS},
      year={2023},
}

@article{wei2022chain,
  title={Chain-of-thought prompting elicits reasoning in large language models},
  author={Wei, Jason and Wang, Xuezhi and Schuurmans, Dale and Bosma, Maarten and Xia, Fei and Chi, Ed and Le, Quoc V and Zhou, Denny and others},
  journal={Advances in neural information processing systems},
  volume={35},
  pages={24824--24837},
  year={2022}
}

@article{chen2017rethinking,
  title={Rethinking atrous convolution for semantic image segmentation},
  author={Chen, Liang-Chieh and Papandreou, George and Schroff, Florian and Adam, Hartwig},
  journal={arXiv preprint arXiv:1706.05587},
  year={2017}
}

@article{xie2021segformer,
  title={SegFormer: Simple and efficient design for semantic segmentation with transformers},
  author={Xie, Enze and Wang, Wenhai and Yu, Zhiding and Anandkumar, Anima and Alvarez, Jose M and Luo, Ping},
  journal={Advances in neural information processing systems},
  volume={34},
  pages={12077--12090},
  year={2021}
}

@article{meyer2025ultrasam,
  title={Ultrasam: a foundation model for ultrasound using large open-access segmentation datasets},
  author={Meyer, Adrien and Murali, Aditya and Zarin, Farahdiba and Mutter, Didier and Padoy, Nicolas},
  journal={International Journal of Computer Assisted Radiology and Surgery},
  pages={1--10},
  year={2025},
  publisher={Springer}
}

@inproceedings{cheng2022masked,
  title={Masked-attention mask transformer for universal image segmentation},
  author={Cheng, Bowen and Misra, Ishan and Schwing, Alexander G and Kirillov, Alexander and Girdhar, Rohit},
  booktitle={Proceedings of the IEEE/CVF conference on computer vision and pattern recognition},
  pages={1290--1299},
  year={2022}
}

@article{yuan2025sa2va,
  title={Sa2va: Marrying sam2 with llava for dense grounded understanding of images and videos},
  author={Yuan, Haobo and Li, Xiangtai and Zhang, Tao and Huang, Zilong and Xu, Shilin and Ji, Shunping and Tong, Yunhai and Qi, Lu and Feng, Jiashi and Yang, Ming-Hsuan},
  journal={arXiv preprint arXiv:2501.04001},
  year={2025}
}

@inproceedings{sun2019deep,
  title={Deep high-resolution representation learning for human pose estimation},
  author={Sun, Ke and Xiao, Bin and Liu, Dong and Wang, Jingdong},
  booktitle={Proceedings of the IEEE/CVF conference on computer vision and pattern recognition},
  pages={5693--5703},
  year={2019}
}

@article{yang2022gatortron,
  title={Gatortron: A large clinical language model to unlock patient information from unstructured electronic health records},
  author={Yang, Xi and Chen, Aokun and PourNejatian, Nima and Shin, Hoo Chang and Smith, Kaleb E and Parisien, Christopher and Compas, Colin and Martin, Cheryl and Flores, Mona G and Zhang, Ying and others},
  journal={arXiv preprint arXiv:2203.03540},
  year={2022}
}

@article{li2023llava,
  title={Llava-med: Training a large language-and-vision assistant for biomedicine in one day},
  author={Li, Chunyuan and Wong, Cliff and Zhang, Sheng and Usuyama, Naoto and Liu, Haotian and Yang, Jianwei and Naumann, Tristan and Poon, Hoifung and Gao, Jianfeng},
  journal={Advances in Neural Information Processing Systems},
  volume={36},
  pages={28541--28564},
  year={2023}
}

@article{wu2025towards,
  title={Towards generalist foundation model for radiology by leveraging web-scale 2d\&3d medical data},
  author={Wu, Chaoyi and Zhang, Xiaoman and Zhang, Ya and Hui, Hui and Wang, Yanfeng and Xie, Weidi},
  journal={Nature Communications},
  volume={16},
  number={1},
  pages={7866},
  year={2025},
  publisher={Nature Publishing Group UK London}
}

@article{weng2025dolphin,
  title={Dolphin v1. 0 Technical Report},
  author={Weng, Taohan and Yan, Chaoran and Liu, Siya and Liu, Xiaoyang and Wu, Yalun and Wang, Boyang and Wang, Boyan and Ren, Jiren and Yan, Kaiwen and Yu, Jinze and others},
  journal={arXiv preprint arXiv:2509.25748},
  year={2025}
}

@article{zhang2025fully,
  title={A Fully Open and Generalizable Foundation Model for Ultrasound Clinical Applications},
  author={Zhang, Hongyuan and Wu, Yuheng and Zhao, Mingyang and Chen, Zhiwei and Li, Rebecca and Zhu, Fei and Zhao, Haohan and Yuan, Xiaohua and Yang, Meng and Qiu, Chunli and others},
  journal={arXiv preprint arXiv:2509.11752},
  year={2025}
}

@inproceedings{zhang2023huatuogpt,
  title={Huatuogpt, towards taming language model to be a doctor},
  author={Zhang, Hongbo and Chen, Junying and Jiang, Feng and Yu, Fei and Chen, Zhihong and Chen, Guiming and Li, Jianquan and Wu, Xiangbo and Zhiyi, Zhang and Xiao, Qingying and others},
  booktitle={Findings of the Association for Computational Linguistics: EMNLP 2023},
  pages={10859--10885},
  year={2023}
}

@article{chen2024huatuogpt,
  title={Huatuogpt-vision, towards injecting medical visual knowledge into multimodal llms at scale},
  author={Chen, Junying and Gui, Chi and Ouyang, Ruyi and Gao, Anningzhe and Chen, Shunian and Chen, Guiming Hardy and Wang, Xidong and Zhang, Ruifei and Cai, Zhenyang and Ji, Ke and others},
  journal={arXiv preprint arXiv:2406.19280},
  year={2024}
}

@article{chen2024huatuogpto1,
  title={Huatuogpt-o1, towards medical complex reasoning with llms},
  author={Chen, Junying and Cai, Zhenyang and Ji, Ke and Wang, Xidong and Liu, Wanlong and Wang, Rongsheng and Hou, Jianye and Wang, Benyou},
  journal={arXiv preprint arXiv:2412.18925},
  year={2024}
}

@inproceedings{kirillov2023segment,
  title={Segment anything},
  author={Kirillov, Alexander and Mintun, Eric and Ravi, Nikhila and Mao, Hanzi and Rolland, Chloe and Gustafson, Laura and Xiao, Tete and Whitehead, Spencer and Berg, Alexander C and Lo, Wan-Yen and others},
  booktitle={Proceedings of the IEEE/CVF international conference on computer vision},
  pages={4015--4026},
  year={2023}
}

@article{ma2024segment,
  title={Segment anything in medical images},
  author={Ma, Jun and He, Yuting and Li, Feifei and Han, Lin and You, Chenyu and Wang, Bo},
  journal={Nature Communications},
  volume={15},
  number={1},
  pages={654},
  year={2024},
  publisher={Nature Publishing Group UK London}
}

@misc{openai2025gpt-5-system-card,
      title={GPT-5 System Card}, 
      author={OpenAI},
      year={2025},
      url={https://openai.com/index/gpt-5-system-card/}, 
}

@misc{anthropic2025claude-sonnet-4-5,
      title={System Card: Claude Sonnet 4.5}, 
      author={Anthropic},
      year={2025},
      url={https://assets.anthropic.com/m/12f214efcc2f457a/original/Claude-Sonnet-4-5-System-Card.pdf}, 
}

@misc{bai2025qwen3vl,
      title={{Qwen3-VL} Technical Report},
      author={Bai, Shuai and Cai, Yuxuan and Chen, Ruizhe and Chen, Keqin and Chen, Xionghui and Cheng, Zesen and Deng, Lianghao and Ding, Wei and Gao, Chang and Ge, Chunjiang and Ge, Wenbin and Guo, Zhifang and Huang, Qidong and Huang, Jie and Huang, Fei and Hui, Binyuan and Jiang, Shutong and Li, Zhaohai and Li, Mingsheng and Li, Mei and Li, Kaixin and Lin, Zicheng and Lin, Junyang and Liu, Xuejing and Liu, Jiawei and Liu, Chenglong and Liu, Yang and Liu, Dayiheng and Liu, Shixuan and Lu, Dunjie and Luo, Ruilin and Lv, Chenxu and Men, Rui and Meng, Lingchen and Ren, Xuancheng and Ren, Xingzhang and Song, Sibo and Sun, Yuchong and Tang, Jun and Tu, Jianhong and Wan, Jianqiang and Wang, Peng and Wang, Pengfei and Wang, Qiuyue and Wang, Yuxuan and Xie, Tianbao and Xu, Yiheng and Xu, Haiyang and Xu, Jin and Yang, Zhibo and Yang, Mingkun and Yang, Jianxin and Yang, An and Yu, Bowen and Zhang, Fei and Zhang, Hang and Zhang, Xi and Zheng, Bo and Zhong, Humen and Zhou, Jingren and Zhou, Fan and Zhou, Jing and Zhu, Yuanzhi and Zhu, Ke},
      year={2025},
      eprint={2511.21631},
      archivePrefix={arXiv},
      primaryClass={cs.CV},
      url={https://arxiv.org/abs/2511.21631}
}

@article{comanici2025gemini,
  title={Gemini 2.5: Pushing the frontier with advanced reasoning, multimodality, long context, and next generation agentic capabilities},
  author={Comanici, Gheorghe and Bieber, Eric and Schaekermann, Mike and Pasupat, Ice and Sachdeva, Noveen and Dhillon, Inderjit and Blistein, Marcel and Ram, Ori and Zhang, Dan and Rosen, Evan and others},
  journal={arXiv preprint arXiv:2507.06261},
  year={2025}
}

@article{wang2025internvl3,
  title={Internvl3. 5: Advancing open-source multimodal models in versatility, reasoning, and efficiency},
  author={Wang, Weiyun and Gao, Zhangwei and Gu, Lixin and Pu, Hengjun and Cui, Long and Wei, Xingguang and Liu, Zhaoyang and Jing, Linglin and Ye, Shenglong and Shao, Jie and others},
  journal={arXiv preprint arXiv:2508.18265},
  year={2025}
}

@inproceedings{pan2025medvlm,
  title={Medvlm-r1: Incentivizing medical reasoning capability of vision-language models (vlms) via reinforcement learning},
  author={Pan, Jiazhen and Liu, Che and Wu, Junde and Liu, Fenglin and Zhu, Jiayuan and Li, Hongwei Bran and Chen, Chen and Ouyang, Cheng and Rueckert, Daniel},
  booktitle={International Conference on Medical Image Computing and Computer-Assisted Intervention},
  pages={337--347},
  year={2025},
  organization={Springer}
}

@article{wang2025citrus,
  title={Citrus-V: Advancing Medical Foundation Models with Unified Medical Image Grounding for Clinical Reasoning},
  author={Wang, Guoxin and Zhao, Jun and Liu, Xinyi and Liu, Yanbo and Cao, Xuyang and Li, Chao and Liu, Zhuoyun and Sun, Qintian and Zhou, Fangru and Xing, Haoqiang and others},
  journal={arXiv preprint arXiv:2509.19090},
  year={2025}
}

@article{ren2016faster,
  title={Faster R-CNN: Towards real-time object detection with region proposal networks},
  author={Ren, Shaoqing and He, Kaiming and Girshick, Ross and Sun, Jian},
  journal={IEEE transactions on pattern analysis and machine intelligence},
  volume={39},
  number={6},
  pages={1137--1149},
  year={2016},
  publisher={IEEE}
}

@inproceedings{carion2020end,
  title={End-to-end object detection with transformers},
  author={Carion, Nicolas and Massa, Francisco and Synnaeve, Gabriel and Usunier, Nicolas and Kirillov, Alexander and Zagoruyko, Sergey},
  booktitle={European conference on computer vision},
  pages={213--229},
  year={2020},
  organization={Springer}
}

@article{zhu2020deformable,
  title={Deformable detr: Deformable transformers for end-to-end object detection},
  author={Zhu, Xizhou and Su, Weijie and Lu, Lewei and Li, Bin and Wang, Xiaogang and Dai, Jifeng},
  journal={arXiv preprint arXiv:2010.04159},
  year={2020}
}

@article{wu2024visionllm,
  title={Visionllm v2: An end-to-end generalist multimodal large language model for hundreds of vision-language tasks},
  author={Wu, Jiannan and Zhong, Muyan and Xing, Sen and Lai, Zeqiang and Liu, Zhaoyang and Chen, Zhe and Wang, Wenhai and Zhu, Xizhou and Lu, Lewei and Lu, Tong and others},
  journal={Advances in Neural Information Processing Systems},
  volume={37},
  pages={69925--69975},
  year={2024}
}

@inproceedings{lai2024lisa,
  title={Lisa: Reasoning segmentation via large language model},
  author={Lai, Xin and Tian, Zhuotao and Chen, Yukang and Li, Yanwei and Yuan, Yuhui and Liu, Shu and Jia, Jiaya},
  booktitle={Proceedings of the IEEE/CVF Conference on Computer Vision and Pattern Recognition},
  pages={9579--9589},
  year={2024}
}

@inproceedings{rasheed2024glamm,
  title={Glamm: Pixel grounding large multimodal model},
  author={Rasheed, Hanoona and Maaz, Muhammad and Shaji, Sahal and Shaker, Abdelrahman and Khan, Salman and Cholakkal, Hisham and Anwer, Rao M and Xing, Eric and Yang, Ming-Hsuan and Khan, Fahad S},
  booktitle={Proceedings of the IEEE/CVF Conference on Computer Vision and Pattern Recognition},
  pages={13009--13018},
  year={2024}
}

@inproceedings{ren2024pixellm,
  title={Pixellm: Pixel reasoning with large multimodal model},
  author={Ren, Zhongwei and Huang, Zhicheng and Wei, Yunchao and Zhao, Yao and Fu, Dongmei and Feng, Jiashi and Jin, Xiaojie},
  booktitle={Proceedings of the IEEE/CVF Conference on Computer Vision and Pattern Recognition},
  pages={26374--26383},
  year={2024}
}

@article{wang2023visionllm,
  title={Visionllm: Large language model is also an open-ended decoder for vision-centric tasks},
  author={Wang, Wenhai and Chen, Zhe and Chen, Xiaokang and Wu, Jiannan and Zhu, Xizhou and Zeng, Gang and Luo, Ping and Lu, Tong and Zhou, Jie and Qiao, Yu and others},
  journal={Advances in Neural Information Processing Systems},
  volume={36},
  pages={61501--61513},
  year={2023}
}

@inproceedings{wang2024git,
  title={Git: Towards generalist vision transformer through universal language interface},
  author={Wang, Haiyang and Tang, Hao and Jiang, Li and Shi, Shaoshuai and Naeem, Muhammad Ferjad and Li, Hongsheng and Schiele, Bernt and Wang, Liwei},
  booktitle={European Conference on Computer Vision},
  pages={55--73},
  year={2024},
  organization={Springer}
}

@inproceedings{lantext4seg,
  title={Text4Seg: Reimagining Image Segmentation as Text Generation},
  author={Lan, Mengcheng and Chen, Chaofeng and Zhou, Yue and Xu, Jiaxing and Ke, Yiping and Wang, Xinjiang and Feng, Litong and Zhang, Wayne},
  booktitle={The Thirteenth International Conference on Learning Representations}
}

@article{gottdiener2004american,
  title={American Society of Echocardiography recommendations for use of echocardiography in clinical trials: a report from the american society of echocardiography's guidelines and standards committee and the task force on echocardiography in clinical trials},
  author={Gottdiener, John S and Bednarz, James and Devereux, Richard and Gardin, Julius and Klein, Allan and Manning, Warren J and Morehead, Annitta and Kitzman, Dalane and Oh, Jae and Quinones, Miguel and others},
  journal={Journal of the American Society of Echocardiography},
  volume={17},
  number={10},
  pages={1086--1119},
  year={2004},
  publisher={Elsevier}
}

@article{wang2024qwen2,
  title={Qwen2-vl: Enhancing vision-language model's perception of the world at any resolution},
  author={Wang, Peng and Bai, Shuai and Tan, Sinan and Wang, Shijie and Fan, Zhihao and Bai, Jinze and Chen, Keqin and Liu, Xuejing and Wang, Jialin and Ge, Wenbin and others},
  journal={arXiv preprint arXiv:2409.12191},
  year={2024}
}

@inproceedings{he2016deep,
  title={Deep residual learning for image recognition},
  author={He, Kaiming and Zhang, Xiangyu and Ren, Shaoqing and Sun, Jian},
  booktitle={Proceedings of the IEEE conference on computer vision and pattern recognition},
  pages={770--778},
  year={2016}
}

@inproceedings{liu2021swin,
  title={Swin transformer: Hierarchical vision transformer using shifted windows},
  author={Liu, Ze and Lin, Yutong and Cao, Yue and Hu, Han and Wei, Yixuan and Zhang, Zheng and Lin, Stephen and Guo, Baining},
  booktitle={Proceedings of the IEEE/CVF international conference on computer vision},
  pages={10012--10022},
  year={2021}
}

@article{dosovitskiy2020image,
  title={An image is worth 16x16 words: Transformers for image recognition at scale},
  author={Dosovitskiy, Alexey},
  journal={arXiv preprint arXiv:2010.11929},
  year={2020}
}

@article{yang2025qwen3,
  title={Qwen3 technical report},
  author={Yang, An and Li, Anfeng and Yang, Baosong and Zhang, Beichen and Hui, Binyuan and Zheng, Bo and Yu, Bowen and Gao, Chang and Huang, Chengen and Lv, Chenxu and others},
  journal={arXiv preprint arXiv:2505.09388},
  year={2025}
}

@article{seedream2025seedream,
  title={Seedream 4.0: Toward next-generation multimodal image generation},
  author={Seedream, Team and Chen, Yunpeng and Gao, Yu and Gong, Lixue and Guo, Meng and Guo, Qiushan and Guo, Zhiyao and Hou, Xiaoxia and Huang, Weilin and Huang, Yixuan and others},
  journal={arXiv preprint arXiv:2509.20427},
  year={2025}
}

@inproceedings{NEURIPS2022_fbb10d31,
 author = {Xu, Yufei and Zhang, Jing and ZHANG, Qiming and Tao, Dacheng},
 booktitle = {Advances in Neural Information Processing Systems},
 editor = {S. Koyejo and S. Mohamed and A. Agarwal and D. Belgrave and K. Cho and A. Oh},
 pages = {38571--38584},
 publisher = {Curran Associates, Inc.},
 title = {ViTPose: Simple Vision Transformer Baselines for Human Pose Estimation},
 url = {https://proceedings.neurips.cc/paper_files/paper/2022/file/fbb10d319d44f8c3b4720873e4177c65-Paper-Conference.pdf},
 volume = {35},
 year = {2022}
}

@inproceedings{DBLP:conf/eccv/LiYLZWWYX22,
  author       = {Yanjie Li and
                  Sen Yang and
                  Peidong Liu and
                  Shoukui Zhang and
                  Yunxiao Wang and
                  Zhicheng Wang and
                  Wankou Yang and
                  Shu{-}Tao Xia},
  editor       = {Shai Avidan and
                  Gabriel J. Brostow and
                  Moustapha Ciss{\'{e}} and
                  Giovanni Maria Farinella and
                  Tal Hassner},
  title        = {SimCC: {A} Simple Coordinate Classification Perspective for Human
                  Pose Estimation},
  booktitle    = {Computer Vision - {ECCV} 2022 - 17th European Conference, Tel Aviv,
                  Israel, October 23-27, 2022, Proceedings, Part {VI}},
  series       = {Lecture Notes in Computer Science},
  volume       = {13666},
  pages        = {89--106},
  publisher    = {Springer},
  year         = {2022},
  url          = {https://doi.org/10.1007/978-3-031-20068-7\_6},
  doi          = {10.1007/978-3-031-20068-7\_6},
  bibsource    = {dblp computer science bibliography, https://dblp.org}
}
\bibliographystyle{colm2024_conference}
\clearpage

\end{document}